\documentclass[preprint,12pt]{elsarticle}

\usepackage{amssymb}
\usepackage{lipsum}
\usepackage{graphicx}
\usepackage{amsmath}
\usepackage{booktabs}
\usepackage{geometry}
\usepackage{tabularx}
\usepackage{url}
\usepackage{hyperref}
\usepackage{colortbl}
\usepackage{caption}
\usepackage{subcaption}
\makeatletter
\renewcommand{\fnum@figure}{Fig. \thefigure}
\makeatother
\biboptions{sort&compress}

\journal{Additive Manufacturing}

\begin{document}

\begin{frontmatter}


\title{MeltpoolNet: Melt pool Characteristic Prediction in Metal Additive Manufacturing Using Machine Learning}


\author[inst1]{Parand Akbari}

\affiliation[inst1]{organization={Department of Mechanical Engineering, Carnegie Mellon University},
            city={Pittsburgh},
            postcode={15213}, 
            state={PA},
            country={USA}}

\author[inst1]{Francis Ogoke}

\author[inst2]{Ning-Yu Kao}
\author[inst1]{Kazem Meidani}

\author[inst2]{Chun-Yu Yeh}
\author[inst1]{William Lee}

\author[inst1,inst2,inst3]{Amir Barati Farimani }

\affiliation[inst2]{organization={Department of Chemical Engineering, Carnegie Mellon University},
            city={Pittsburgh},
            postcode={15213}, 
            state={PA},
            country={USA}}
\affiliation[inst3]{organization={Machine Learning Department, Carnegie Mellon University},
            city={Pittsburgh},
            postcode={15213}, 
            state={PA},
            country={USA}}

\begin{abstract}
Characterizing meltpool shape and geometry is essential in metal Additive Manufacturing (MAM) to control the printing process and avoid defects. Predicting meltpool flaws based on process parameters and powder material is difficult due to the complex nature of MAM process. Machine learning (ML) techniques can be useful in connecting process parameters to the type of flaws in the meltpool. In this work, we introduced a comprehensive framework for benchmarking ML for melt pool characterization.  An extensive experimental dataset has been collected from more than 80 MAM articles containing MAM processing conditions, materials, meltpool dimensions, meltpool modes and flaw types. We introduced physics-aware MAM featurization, versatile ML models, and evaluation metrics to create a comprehensive learning framework for meltpool defect and geometry prediction. This benchmark can serve as a basis for melt pool control and process optimization. In addition, data-driven explicit models have been identified to estimate meltpool geometry from process parameters and material properties which outperform Rosenthal estimation for meltpool geometry while maintaining interpretability. 

\end{abstract}

\begin{keyword}
Additive Manufacturing \sep Machine Learning \sep Melt pool \sep Process Map

\end{keyword}

\end{frontmatter}



\section{Introduction}
\label{sec:sample1}

Metal additive manufacturing (MAM), as a new and disruptive manufacturing technology, has paved the way toward the next industrial revolution by enabling complex component manufacturing \cite{BERMAN2012155}. MAM is able to fabricate parts and components with complex geometries with special alloys and materials \cite{WANG2020101538,JOHNSON2020101641}. 
Compared to conventional manufacturing technologies such as milling and casting, it has several benefits, including reduced material consumption, shorter lead time, higher performance and reliability, and the capability to manufacture complex geometries\cite{jiang2020path, debroy2018additive}. Due to these special characteristics, it has been commercialized for use in high-tech industries such as the aerospace, automotive, energy, and biomedical fields\cite{wang2016topological, liu2021review, zhakeyev2017additive, leal2017additive}. Although additive manufacturing has attracted a great deal of scientific and engineering work in industry and academia, increasing the production speed, scale, and quality of printed products is still a major challenge \cite{BERMAN2012155,WANG2020101538}.  Specifically, the quality of printed parts is a major issue, as defects negatively influence the structural integrity of the part. It is also the most challenging one to overcome due to the complex, multi-scale physics involved in the AM process\cite{markl2016multiscale}.


\vspace{3mm}

To determine the optimal processing window for part production with minimal defects, various experimental monitoring techniques are often implemented. These methods, are divided into ex-situ and in-situ monitoring techniques, based on whether it is conducted after the part has been produced or while the component is being manufactured, respectively \cite{BAYAT2021102278}. These monitoring techniques can be employed to gain further insight into the generation mechanism of these defects. Most  monitoring and control techniques focus on the characteristics of the melt pool, since common defects in MAM, i.e. lack of fusion, keyhole, and balling are rooted in melt pool and melt pool dynamics \cite{SCIME2019151}. The melt pool is a locally melted zone generated by short-time laser or electron beam irradiation on powder particles, and it is pivotal to the properties of the overall product. As a result, melt pool monitoring is often used to ensure the quality of additively manufactured parts \cite{Dilip2017, Yu2016,CRIALES201722,KESHAVARZKERMANI201983}. To find an optimal processing window to achieve a desired melt pool geometry (depth, width, and length), processing parameters such as beam power, scanning speed, hatch spacing, layer thickness, and beam diameter must be selected carefully, resulting in rapidly and reliably produced parts with desired properties, high density, and low porosity \cite{gordon2020defect}.

\vspace{3mm}

However, due to the complex multi-physics and multi-scale nature of AM processes, as well as the significant influence of processing parameters on the quality of the printed products, AM has undergone a paradigm shift from purely physics based approaches to physics-based and data-driven approaches\cite{wang2020machine}. Therefore, data-driven analysis and machine learning (ML) has become a routine step in advanced manufacturing applications, and increasingly popular in AM research \cite{JOHNSON2020101641}. In addition, experimental monitoring techniques can be costly,  inefficient, and often requires extensive preparation and arduous calibration. Alternatively, applying ML models that are built upon experimental data can provide a more cost-effective solution. With a reliable training data set, the trained machine learning models can make accurate predictions, and determine the optimal processing parameters in an efficient manner. Therefore, utilizing ML methods in additive manufacturing improves product quality, optimizes manufacturing processing parameters, and reduces cost \cite{Meng2020}.  Furthermore, in comparison with the experimental techniques that are solely aimed and designed for only a specific subject, ML models are much more flexible and can be easily adjusted upon the application \cite{BAYAT2021102278}.

\vspace{3mm}

However, creating machine learning models for metal additive manufacturing requires overcoming challenges related to the limited amount of data. The available data on AM is highly variable and expensive to gather, and thus much smaller than those available for other machine learning tasks. Additionally, this data often has large heterogeneity in input processing parameters, and therefore the appropriate selection of learning algorithms is necessary. Moreover, additive manufacturing processes require knowledge about how process parameters affect material properties and melt pool dynamics to produce defect-free parts. Thus, creating machine learning models in additive manufacturing is difficult, due to the lack of sufficient data and the complexity of MAM processes \cite{JOHNSON2020101641}.

\vspace{3mm}

Despite these challenges, researchers have employed ML in laser powder bed fusion \cite{OGOKE2021102033, SCIME2019151, LEEETAL2019}. For instance, Tapia et al. built a surrogate model for laser powder bed fusion of 316L stainless steel to predict the melt pool depth with an MAE of 10.91 $\mu m$ of 96 single-track prints from the laser power, velocity, and beam diameter and associated simulations \cite{TAPIAETAL2018}. Scime et al. used machine learning models to detect keyholing porosity and balling instabilities for laser powder bed fusion of Inconel 718 \cite{SCIME2019151}. Lee et al. created a data analytics framework for determining melt pool geometry with an $R^2$ correlation of 0.94 in powder bed fusion process of single tracks of Inconel 718 and Inconel 625 \cite{LEEETAL2019}. Yuan et al. employed a machine learning based monitoring of laser powder bed fusion of 316L stainless steel based on  in-situ laser single-track videos, enabling the prediction of laser powder bed fusion track widths with an $R^2$ of 0.93 \cite{YUANETAL2018}. Gaikwad et al. developed machine learning based predictive models of single tracks of 316L stainless steel using pyrometer and high-speed video camera data \cite{GAIKWAD2020101659}. Their best ML prediction for meltpool width is achieved by a sequential decision analysis neural network (SeDANN) model with an $R^2$ of 0.87. While these papers have been successful in using machine learning to enhance metal additive manufacturing applications, their experimental dataset, processing parameters, and material used are limited. Our benchmark is built upon a broader source of data, which allows us to optimize and control part quality on a larger range of processing parameters and materials. Our dataset comprises of at least 80 sources of experimental data \cite{app11072962, Jakumeit_2020, Chen_2020, Klassen_2014, jmmp3010021, Cunningham849, ma12101706, met8070475, met10091222, Kiss, CHEN2021129205, ANDREOTTA201736, ZHANG2019297, ZHANG2021109501, SHI20169, NAYAK2020106016, LI2021298, GARGALIS2021117130, SONG2019551, LU2018801, HEELING2017116, ANDREAU201921, GRANGE2021116897, GUO2021117000, SHUAI2019606,KARIMI201988, ROEHLING2017197, KOEPF2018119, JOHNSON2019199, VRANCKEN2020464, MUKHERJEE2018442, MIRKOOHI2019532, LADANI201713, GUO2019600, CALTA2020101084, HYER2020101123, STOPYRA2020101270, PANKAJ2019, Cheng2013, Piazza2020, Imani2020, Hanemann2020, Zielinski2020, Cullom2020, Francis2018, Riedlbauer2017, Reisgen2020, Obidigbo2017, Giordimaina2017, Kamath2016, Chen2020, NAKAPKIN2019, Martin2019, Rosser2019, Lu2020}, obtained from experiments each of which is conducted on a limited number of alloys with one AM process. With this work, we unify these disparate data sources into a larger repository for melt pool characterization under a wide range of processing parameters and alloys.

\begin{figure}\centering
\includegraphics[scale=0.8]{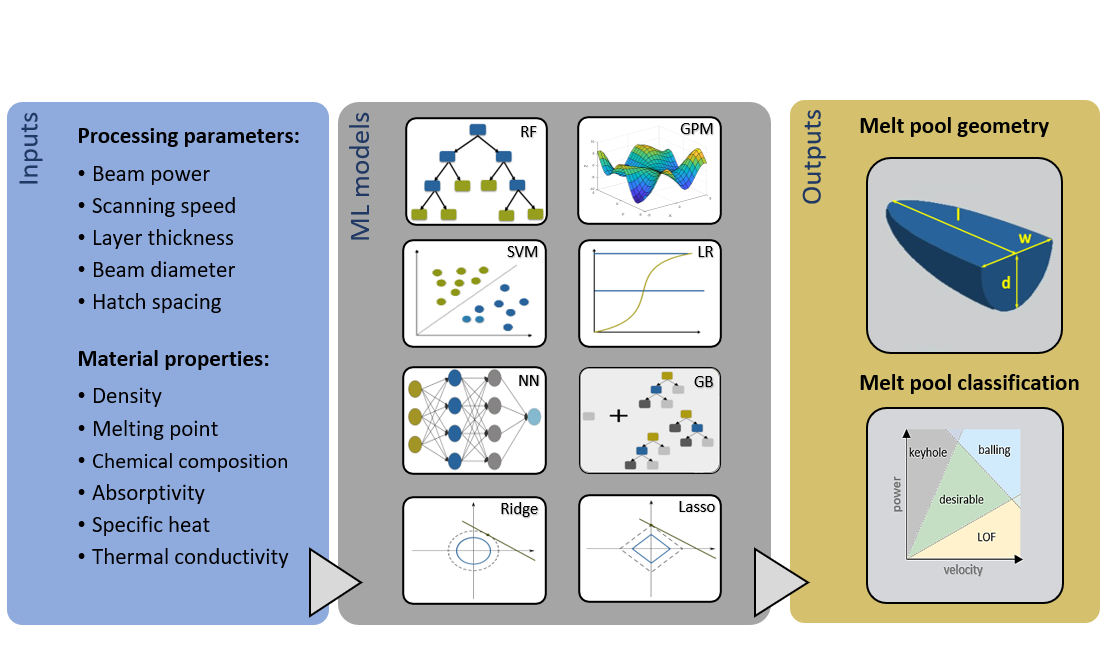}
\caption{Features, ML models, and tasks implemented in our MeltpoolNet benchmark. }
\label{fig:fig1}
\end{figure}

\vspace{3mm}

In this work, we aim to characterize  melt pool behavior (predict its geometry and the types of flaws) by developing a suite of additive manufacturing machine learning methods, called MeltpoolNet. MeltpoolNet uses the comprehensive experimental dataset collected by authors to build ML models. MeltpoolNet will enable the optimization of processing parameters for a desired melt pool, as well as the prediction of porosity in the printed parts. Our dataset contains experimental data with varying processing parameters, materials and MAM process types (PBF and DED). We also assess the impact of various parameters associated with the build process on the performance of these models. Additionally, a data-driven model identification method, with higher interpretability in comparison with ML models, is developed to find the explicit relations between the dataset processing parameters and material properties.

\begin{figure}\centering
\includegraphics[scale=0.11]{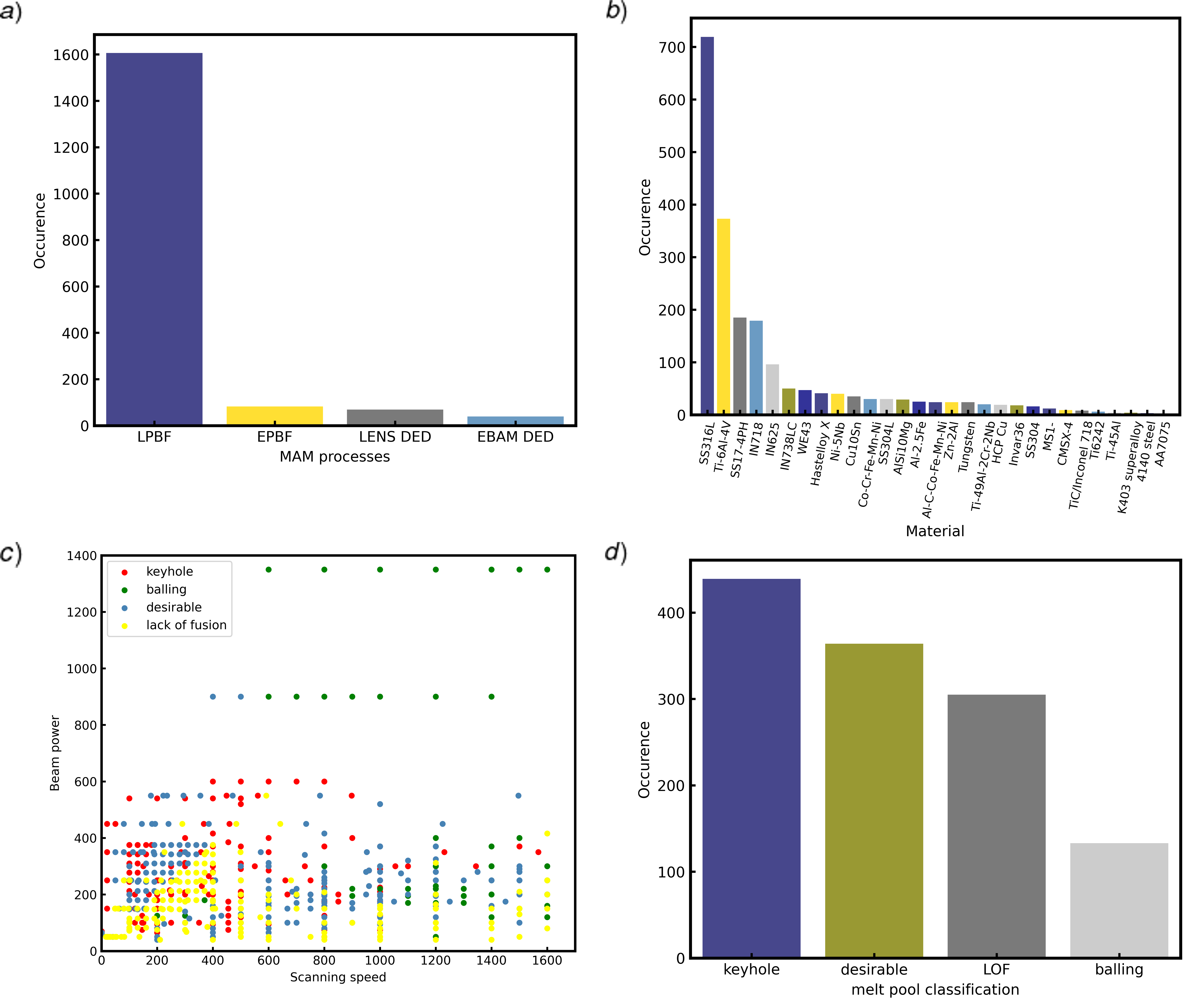}
\caption{a) Distribution of different MAM processes in our benchmark dataset, b) Materials studied in our benchmark, and their occurrence, c) A presentation of our dataset for classification task for different materials in L-PBF process, d) Melt pool flaw and mode type distribution in our dataset. }
\label{fig:fig2}
\end{figure}

\section{Methodology} 
Fig.\ref{fig:fig1} shows the MeltpoolNet framework, containing the raw features of the dataset, the featurization, the implemented ML models and the associated tasks. In this section, we discuss the dataset collection and curation, feature engineering, and ML algorithms.

\vspace{3mm} 
\subsection{\normalfont{Data collection }}
The data on both the melt pool geometry and flaw and mode type were collected from published work in manufacturing and materials journals, specifically those which report experimental data for these properties. Primarily, data was extracted from the figures and tables of these papers. In order to accurately extract data from plots and figures, the Plot Digitizer program was used \cite{plotdigitizer}. The processing parameters and material properties used in each experiment are also gathered to be used as input to our ML models.  


\begin{figure}\centering
\includegraphics[scale=0.11]{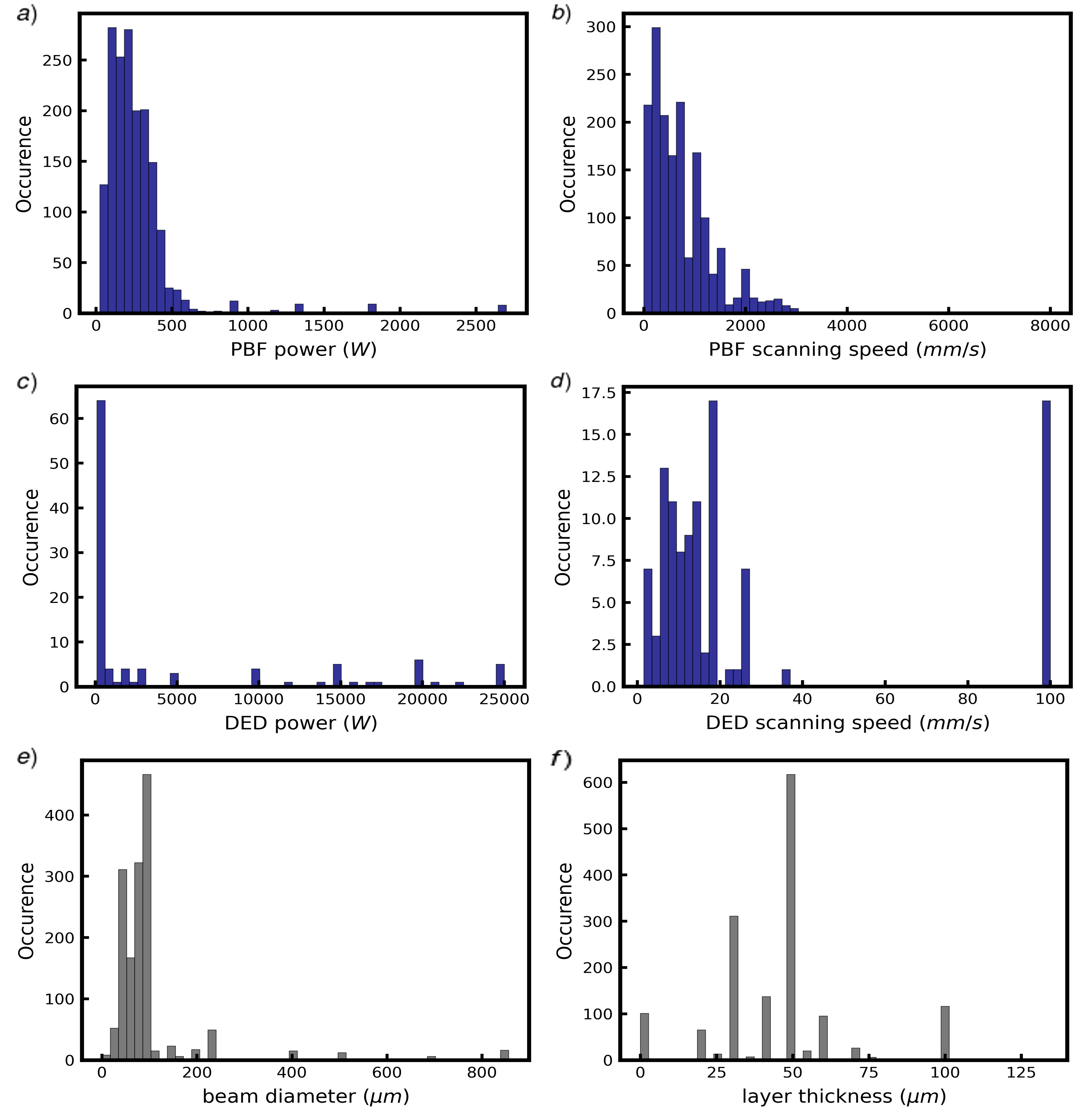}
\caption{Processing parameters histograms and their distribution from our datasets. A) PBF power histogram and occurence, b) PBF scanning speed histogram and occurrence  c) DED power histogram and occurence, d) DED scanning speed histogram and occurence
e)Beam diameter histogram d) PBF Powder layer thickness histogram }
\label{fig:fig3}
\end{figure}
\vspace{3mm} 
\subsection{\normalfont{Datasets}}

MeltpoolNet is built upon data obtained from literature. Our current dataset offers about 2200 data points. Each data point contains the processing parameters and material properties as input features, and the geometry of the meltpool (Width, Length and Depth) along with the type of the defects or meltpool modes (Lack of Fusion, Keyhole, Balling, Desirable) as labels.
\vspace{3mm}

Based on the beam input energy density, calculated based on a combination of beam power, scanning speed, and diameter, either a conduction mode or keyhole mode melt pool is expected.\cite{KING20142915} The melt pool in its conduction mode with semi-circular shape without any defects and porosities is termed a desirable meltpool. However, in high energy density cases, the melt pool depth becomes larger than half of the melt pool width. In this case, a semi-circular meltpool shape cannot be assumed, and the meltpool can be considered as keyhole mode. It is worth mentioning that keyhole mode will not always lead to keyhole porosity. \cite{KING20142915} In addition, lack of fusion defects occur when the melt pool depth or the overlap depth between two adjacent melt tracks is lower than the powder layer thickness. Balling phenomena also happen in high beam power and scanning speed with high melt pool length to width ratio, resulting in instability or in high scanning speed and low beam power where there is not enough bonding to the substrate. 
\vspace{3mm}

These data points have been collected from experiments conducted using various alloys, AM processes (PBF and DED), and processing parameters. Using the above-mentioned features, the machine learning algorithms are then deployed to predict different properties of the melt pool. These properties are being predicted in different regression and classification tasks. The regression task predicts the depth, width, and length of the melt pool, and the classification task predicts the defect mode of the meltpool. 
Metal AM processes can be categorized based on the heat source (laser or electron beam), and the way the raw material is supplied (powder or wire feed). Our dataset of AM processes is comprised of two classes of Powder Bed Fusion (PBF) - electron beam powder bed fusion (E-PBF) and laser powder bed fusion (L-PBF), in addition to two Directed Energy Deposition (DED) processes - laser based powder fed AM process (LENS) and electron beam wire fed AM process (EBAM) .(Fig. \ref{fig:fig2}a)

\vspace{3mm}
The regression tasks dataset consist of the all the four above mentioned AM processes, however, the classification dataset includes L-PBF and E-PBF processes. We have investigated the geometry (depth, width, and length) as well as the melt pool defect and mode classification of 29 different alloys in AM processes.(Fig. \ref{fig:fig2}b)  The chemical composition, density, specific heat, thermal conductivity, and melting temperature of the studied materials are presented in Table \ref{sec:sample:appendix:tab:tableA1} and Table \ref{sec:sample:appendix:tab:tableA2}.  The meltpool classification is separated into four different common classes seen during AM build processes, namely desirable (no defect present), keyhole, lack of fusion, and balling (Fig. \ref{fig:fig2}d). Fig. \ref{fig:fig2}c presents our benchmark classification dataset for different alloys under a wide range of processing parameters in L-PBF process.

\vspace{3mm}
Furthermore, the distributions of beam power, scanning speed, and beam diameter in PBF and DED processes studied in our benchmark are shown in Fig. \ref{fig:fig3} a-e. As presented in the Fig. \ref{fig:fig3}, the PBF process utilizes lower beam powers and higher scanning speeds as compared to DED process. In addition, the powder layer thickness distribution in PBF dataset is depicted in Fig. \ref{fig:fig3} f. During the ML training process, only the data points with data present for each desired input feature are chosen to comprise the dataset. For the baseline input features listed in Table 2, this allows for 93$\%$ of the labeled data points to be used in the dataset for the regression task, and 88$\%$ of the data points for the classification task. Our benchmark contains 2200 data samples on the melt pool properties, and is expected to grow in an ongoing manner. As such, we highly welcome contribution from other public data collections.

\begin{table}[ht]
\begin{center}
\caption{Our benchmark dataset details. (Tasks, splitting method, and metrics).}\vspace{2mm}
\label{tab:table1}
\scalebox{0.7}{
\begin{tabular}{c| c c  c c c}
\toprule [1pt]
Category & Dataset            & Tasks     & Rec - split   &   Rec - metric    \\
\midrule [2pt]
Depth of melt pool  & 1400 &  Regression & Random & $R^2$ - MAE \\ [1ex]
Width of melt pool  & 1200  & Regression &  Random & $R^2$ - MAE\\ [1ex]
Length of melt pool  & 320  & Regression & Random & $R^2$ - MAE \\ [1ex]
Melt pool classification & 1200 & Classification & Random & Classification accuracy - AUC-ROC \\ [1ex]
\bottomrule
\end{tabular} 
}
\end{center}
\end{table}   

\vspace{3mm}

\subsection{\normalfont{Featurization}}
During featurization, we select and construct features and feed them as input to our ML models for prediction. To generate a well-trained ML model for a specific property prediction, a sufficient number of features has to be distinguished, especially due to the complexity of the physical phenomena involved in MAM. Since the specific AM processes used for an experiment is a categorical feature, we used one hot encoding to featurize this aspect of the data. One hot encoding converts categorical features, where each value belongs to one of $n$ non-numeric categories, to numeric categories that can be processed by conventional ML models. To do so, it transforms a categorical feature with $n$ possibilities to $n$ binary encoding features. For a data sample that belongs to category $k$, a $1$ will be placed in the $k$-th encoding feature, and $0$ will be placed in the other $n-1$ encoding features. In this manner, the prediction task can be undertaken while acknowledging that different heat sources and the way feed stock materials are supplied will have different relationships with the other features, and the prediction target. 
 \vspace{3mm} 
 
A specific sample collected from literature may not include all of the process parameters and features reported. Therefore, for a given feature combination, points that are missing one or more features are removed from the training and prediction tasks. This is done in order to ensure that consistent information is provided to the model for a given feature set.

\vspace{3mm}
Additionally, for regression tasks, the processing parameters, i.e., laser/electron beam power and scanning speed along with the beam diameter for classification task, and material properties (density, specific heat, thermal conductivity, and melting temperature) are considered to be the baseline features. Table \ref{tab:table2} presents this baseline featurization alongside other features utilized in our benchmark. In addition to featurizing the alloys studied in our benchmark with their thermal properties, we have employed other various methods: material one hot encoding, material chemical composition (wt\%),  and elemental featurization. Furthermore, other processing parameters, such as hatch spacing, absorptivity, and powder layer thickness (for PBF processes) have been added to the baseline featurization to observe their effects on the accuracy of our models. The material elemental featurization and absorptivity featurization are elaborated on in the following sections. 

\begin{table}[ht!]
\begin{center}
\caption{Featurization employed in our benchmark ML models.}\vspace{2mm}
\label{tab:table2}
\scalebox{0.58}
{
\begin{tabular}{c c | c  c c}
\toprule [1pt]
Baseline featurizarion & & Other features && \\ [2ex]
\midrule [2pt]

Processing parameters & Material properties & Materials featurization &   Other processing parameters \\ [2ex]
\midrule [1pt]

Process one hot encoding &   Density & One hot encoding &  Hatch spacing \\ [2ex]

Beam power & Specific heat& Chemical composition (wt\%) & Absorptivity \\ [2ex]
 
Beam scanning speed &Thermal conductivity & Elemental featurization  & Powder layer thickness (PBF process) \\ [2ex]
 
Beam diameter (classification task) & Melting temperature && \\ [2ex]
 
\bottomrule
\end{tabular}
}
\end{center}
\end{table}

\subsubsection{\normalfont Elemental Features}

In metal additive manufacturing, alloys are often used as the raw material to print the product, containing multiple elements with specific concentrations. Therefore, we integrate these elemental properties into our feature set used to predict the melt pool properties.

\vspace{3mm}
The element properties to be considered in our machine learning models are listed in Table \ref{tab:table3}. To access these properties, we make use of the Mendeleev package \cite{mendeleev}, which contains well-documented elemental properties. The alloys in our dataset are comprised of 19 unique elements, and each alloy contains multiple elemental components. In order to account for the influence of the relative concentration of each element in the alloy in the overall alloy property, we use a linear mixture rule \cite{CARRUTHERS1988351}:
\begin{equation}
x_i = \sum_{j}^{N} x_{ij}  a_j 
\end{equation}
where $x_i$ is the $i^{\text{th}}$ feature, $x_{ij}$ is the $i^{\text{th}}$ feature of the $j^{\text{th}}$ component, and $a_i$ is the $j^{\text{th}}$ element fraction. 

\begin{table}[ht!]
\begin{center}
\caption{Properties utilized in our elemental featurization, a method for featurizing materials.}\vspace{2mm}
\label{tab:table3}
\scalebox{1}
{
\begin{tabular}{c c }
\toprule [1pt]
Elemental featurization&  \\ [1ex]
\midrule [2pt]
Atomic number & Heat of fusion \\ [1ex]
Atomic volume & Electron affinity \\ [1ex]
Ionization energy&   \\ [1ex]
\bottomrule
\end{tabular}
}
\end{center}
\end{table}
\subsubsection{\normalfont Absorptivity coefficients} 

 When a laser or electron beam impacts the metal substrate during metal additive manufacturing, some fraction of the energy is absorbed, while the rest is reflected or lost. The magnitude of the absorbed energy relative to the energy input can be expressed in terms of an absorptivity coefficient, $\eta $. The absorptivity can vary as a function of the melt pool shape, due to reflection effects.  Once the peak temperature is higher than the boiling point of the material, due to the multireflection of the laser beam between the walls of the depression, the keyhole depression deepens, resulting in rising absorbed power \cite{trapp2017situ, simonds2021causal}. In this situation, the absorptivity (called absorptivity 1 in our benchmark), $\eta$, can be expressed as \cite{Ganetal2021}: 
\begin{equation}
\eta = 0.7[1-\exp(-0.6\frac{\eta_m P}{(T_1 - T_0)\pi \rho C_p V r_0^2 })]
\end{equation}
where $\eta_m$ is minimum absorptivity, i.e., absorptivity on flat melt surface,  P the laser power, $T_1$ the melting point of the alloy, $T_0$ the temperature far from the melt pool, $\rho$ is the material density, $C_p$ specific heat of the alloy, V  laser scan speed, and $r_0$ is beam radius. 

In addition, if  the peak temperature is lower, and we have a semicircular shaped melt pool, the absorptivity (called absorptivity 2 in our benchmark) can be expressed as \cite{TANG201739}:
\begin{equation}
\eta = \frac {\pi k (T_1 - T_0) W + e \pi \rho C_p (T_1 - T_0) V W^2 /8}{P} 
\end{equation}
where $k$ is thermal conductivity of the alloy, and $W$ is the width of the melt pool. Calculating the absorption coefficients with regards to above mentioned equations, and adding them as new features to our datasets, increased the accuracy of the implemented models in some tasks significantly. 
\vspace{3mm}

\subsection{\normalfont{Dataset splitting and Metrics}}
For each dataset, a metric and a splitting pattern has been defined that best fits the properties of the dataset. Table \ref{tab:table1} lists the details of datasets in our collection, comprising of tasks, recommended splits and metrics. For regression tasks, the properties are continuous values, and for classification tasks, categorical class labels. Due to the fact that our dataset input parameters are highly heterogeneous, our dataset is then normalized. The equation for normalization utilized in our benchmark is given by:

\begin{equation}
x_{Normalized} = \frac{x - \bar{x}}{\sigma}
\end{equation}
where $\bar{x}$ is the mean of the input parameters, and $\sigma$ is the standard deviation. 

\vspace{3mm}
Machine learning methods require datasets to be split into training/test subsets for model selection to verify the performance of the models on unseen data.  To do so, we split our data set into a training and testing partition, where the models are trained on the training partition, and the test partition is reserved solely for evaluating model performance. 
\vspace{3mm} 
Additionally, $k$-fold cross validation is performed to gain more insight into the model performance by taking into account multiple different training and test partitions. During $k$-fold cross validation, the entire dataset is split into $k$ partitions. Following this splitting process, $k-1$ partitions are used to train the model, and the $k$-th partition is used to test the model. This process is repeated for a given split of $k$ partitions $k$ times, such that every partition is used as the testing partition once.  
\vspace{3mm} 
To estimate the power of our machine learning models on unseen data, we first shuffled our datasets randomly, and then performed 5-fold cross validation. In 5-fold cross validation, we split the data into 5 groups, and  over 5 iterations, we take one group as a test dataset, and the remaining groups as the training dataset. Thus, we then evaluated the accuracy of our models as the averaged accuracy over all five iterations. 

\vspace{3mm}
Our dataset contains both a regression task (geometry of the melt pool), and a classification task (the melt pool defect class). The regression task evaluation metrics used are the mean absolute error (MAE) and the $R^{2}$ coefficient of determination. For the classification task, the evaluation metric is defined as the fraction of correctly predicted samples over the given dataset. The classification task is also evaluated by the area under the receiver operating characteristic curve (AUC-ROC). 

\subsection{\normalfont{Models}}
We tested the performance of various machine learning models on our datasets explained previously. These models will be introduced briefly below, and their performance metrics will be discussed in more detail in the next section. 

\subsubsection{\normalfont Random Forest 'RF'}
The Random Forest (RF) algorithm which is used in both classification and regression tasks, utilizes ensemble learning \cite{RF}. RF models consist of many decision trees trained independently on random sub-samples of the dataset. It combines outputs at the end of the process; for the regression task, it predicts by returning the average of the outputs of the decision trees, and for the classification task, it returns the majority vote of the decision trees. The Random Forest model has been used for both our regression and classification tasks. 

\subsubsection{\normalfont Gaussian Process model ('GPR' and 'GPC')}
Gaussian Process Regressors (GPR) and Gaussian process Classifiers (GPC) are probabilistic algorithms \cite{Rasmussen2004} . This model returns a probability distribution over all possible output values due to its Bayesian approach, and can predict well on small datasets. Gaussian process models have been used for both the regression and classification tasks. 

\subsubsection{\normalfont Support Vector Machine ('SVR' and 'SVC')}
Support Vector Machines (SVMs) are a class of machine learning algorithms used for both regression and classification tasks. For the classification task, it predicts by finding a hyperplane which separates data points of different categories by maximizing the distance, called margin, between the nearest data points and the decision boundary  \cite{svm}. For the regression task, it gives a decision boundary at a distance from the original hyperplane such that the data points closest to the hyperplane, called the support vectors, are within that boundary line. Support vector machine models have also been used for both the regression and classification tasks in this work. 

\subsubsection{\normalfont Ridge Linear Regression 'Ridge'}
Ridge regression is a regularized version of linear regression, used to reduce overfitting which can occur when the model is overspecified for a small dataset.\cite{Ridge} This method performs $L_2$ regularization. When overfitting occurs in a model without regularization, the training error is very small since the complex model can fit the training data points well, but the test error is very large. Ridge regression shrinks the parameters by imposing a constraint on their values, helping to reduce the model complexity and overfitting. Ridge regression model has been used for the regression task. 

\subsubsection{\normalfont Lasso Linear Regression 'Lasso'}
The Least Absolute Shrinkage and Selection Operator (LASSO) is also a regularization method used to modify linear regression for a more accurate prediction with less overfitting.\cite{lasso} Lasso regression performs $L_1$ regularization. Due to the $L_1$ penalty, this model nullifies parameters where needed to reduce the model complexity and overfitting. The Lasso regression model has also been used for the regression task in this work.

\subsubsection{\normalfont Gradient Boosting Trees 'GB'}
Gradient boosting algorithms can be used for both regression and classification tasks. This model also utilizes ensemble learning, and contains multiple decision trees. \cite{GB} In comparison with Random Forest models, Gradient boosting combines results along the process. Additionally, it builds one tree at a time, and each new tree tries to correct errors produced by the previous one. The Gradient boosting model has been used for both the regression and classification tasks. 

\subsubsection{\normalfont Logistic Regression 'LR'}
Logistic regression models are used for classification tasks \cite{LR}.  In Logistic regression, a weighted combination of input features are passed through the sigmoid function, and it predicts a class by comparing the probability values which are the Logistic function outputs, with a  probability threshold. The Logistic regression model has been used in the classification task.

\subsubsection{\normalfont Neural Network 'NN'}
Neural networks are machine learning algorithms that can be used in both regression and classification tasks \cite{NN}.  It is inspired by the human brain, and mimics its operation, with individual neurons exchanging information and modifying connections during training. Neural networks are comprised of an input layer, one or more hidden layers, and an output layer. A fully connected neural network model has been used for both the regression and classification tasks. 

\subsubsection{\normalfont XGBoost}
XGBoost is a supervised learning model that implements gradient boosted trees in order to generate predictions \cite{Chen2016}. In comparison to the vanilla gradient boosted tree described earlier, XGBoost performs tree regularization to decrease the chance of overfitting to the data. XGBoost is used for both the regression and classification tasks. As part of this work, all methods are implemented using the open source scikit-learn package. This package also contains the methods used to split the dataset into its train and test set. It also was used to evaluate the performance of our models with the built in methods that calculated the MAE, $R^2$, AUC-ROC, and accuracy of the models.

\subsubsection{\normalfont Description of Hyperparameter optimization}
The performance of ML models often depends on the hyerparameters selected, which control the details of the model configuration used during training. The hyperparameters controlling the training process should be adjusted by the user before training the model, and model parameters learned during the training process are affected by hyperparameters. Thus, choosing an optimal value for ML algorithms' hyperparameters is of vital importance, since they can significantly influence the prediction performance of the ML models. Therefore, in order to determine the optimal hyperparameters for prediction, the python package Hyperopt was utilized for each of the various  regression and classification methods mentioned above \cite{hyperopt}. Hyperopt implements the Tree-structured Parzen Estimator (TPE) algorithm to optimize an objective function, which in this case was the validation $R^2$ score for the regression task and the validation multiclass logarithmic loss for the classification task. The full set of parameters explored during the search process for meltpool depth regression and classification are listed in Table \ref{tab:table4}. The optimized ML hyperparameters for meltpool width and length are reported in Table \ref{sec:sample:appendix:tab:tableA3}.

\vspace{3mm}

For the Random forest and Gradient Boosting algorithms, the 'n-estimators' hyperparameter represents the number of decision trees in the forest. (Table \ref{tab:table4}) The range of 1 to 500 trees has been investigated to find the optimal value. While a higher number of decision trees improves these models' prediction performance, however, it results in longer computation time.  \cite{probst2019hyperparameters} The hyperparameter optimized for the Gaussian Process based algorithms (GPR, GPC) is the kernel used to define the covariance function between datapoints, defined similarly to the kernels used in SVM models.

\vspace{3mm}

For the SVM algorithm, the kernel and the regularization hyperparameter $C$ have been optimized. Four different SVM kernels, namely, the linear, polynomial, radial basis function (rbf), and sigmoid kernels, were studied. For the linear kernel, SVR fits linear support vectors over the data points, and SVC classifies data points with a linear decision boundary. When the data points are not linearly separable, non-linear kernels- polynomial, rbf, and sigmoid- can perform better, so the SVM kernel trick comes into play. The SVM kernel trick maps the data points into a higher dimension space to make them linearly separable. \cite{scholkopf2002learning} The regularization parameter, $C$, is a penalty parameter for the misclassified data points which controls the trade-off between the number of points being classified correctly, and the distance of the decision boundary to each class. A smaller value of $C$, or a lower penalty, allows for more misclassified data points, while larger values of $C$, or a greater penalty, minimizes the misclassified data points, due to the smaller margin decision boundary. \cite{DUAN200341} For $C$, the range of 1 to 1000 was studied. 

\vspace{3mm}

For the Neural network model, two different hyperparameters are optimized, the number of neurons and the regularization parameter alpha. Five different neuron numbers for each layer- 32, 64, 128, 256, and 512- have been explored. A higher number of neurons increases the model complexity, but may result in overfitting. Thus, an optimal value for the number of neurons in each layer should be obtained. Alpha is a regularization parameter which constrains the $L_2$ norm of the model weights, thereby reducing model complexity and discouraging overfitting. In order to optimize this hyperparameter, the range of 1e-7 to 1e-1 has been investigated. For Logistic regression, the range of 1 to 500 has been studied to find the optimal value for the regularization parameter, $C$, defined as the inverse of the regularization strength. Therefore, smaller values of $C$ leads to stronger regularization. 
\vspace{3mm} 

To ensure that this dataset will be robust to ongoing updates and data additions, the hyperparameters can be automatically selected through the  optimization process discussed above. This will ensure that the optimal model configurations are being used in each learning scenario. A study of the model performance compared to the amount of data samples has been conducted in Figure \ref{fig:fig11}.

\vspace{3mm}

\begin{table}[hbt!]
\begin{center}
\caption{Hyperparameters and their range studied in our benchmark ML models for meltpool depth regression and classification.}\vspace{1mm}
\label{tab:table4}
\scalebox{0.5}{
\begin{tabular}{c c c c c}
\toprule [1pt]
ML Task & {Models} & Hyperparameters & Value &  Range studied   \\ [1ex]
\midrule [2pt]
Depth Regression & RF & n\_estimators &	500 & 1-500	\\ [2ex]
& GPR & kernel & ConstantKernel(1.0, (1e-1, 1e3))*RBF(1.0, (1e-3, 1e3)
	&		RBF, DotProduct, Matern, RationalQuadratic  \\ [1ex]
 & SVR & C & 992	&1-1000	\\ [0.8ex]
 &  & kernel & 'rbf'	& ['linear', 'poly', 'rbf', 'sigmoid']	\\ [2ex]
 & GB & n\_estimators & 489 &1-500	\\ [2ex]
 & NN & number of neurons &(64,32,64)	& [32, 64, 128, 256, 512]\\ [0.8ex]
 &  & alpha & 0.065533	& 1e-7 -  1e-1\\ [2ex]
Classification & RF & n\_estimators &	160 &1-500\\ [2ex]
 & GPC  & kernel & RBF(length\_scale=1)	&\\ [1ex]
 & SVC & C & 868	&1-1000	\\ [0.8ex]
 &  & kernel & 'rbf'	&['linear', 'poly', 'rbf', 'sigmoid'] \\ [2ex]
 & LR & C & 415	& 1-500 \\ [1ex]
 & GB & n\_estimators & 162 &1-500	\\ [2ex]
  &  NN & number of neurons & (32,128,128)	& [32, 64, 128, 256, 512]\\ [0.8ex]
 &  & alpha & 0.020241313	& 1e-7 -  1e-1\\ [1ex]
\bottomrule 
\end{tabular}}
\end{center}
\end{table}

\vspace{3mm}

\begin{figure}\centering
\includegraphics[scale=0.093]{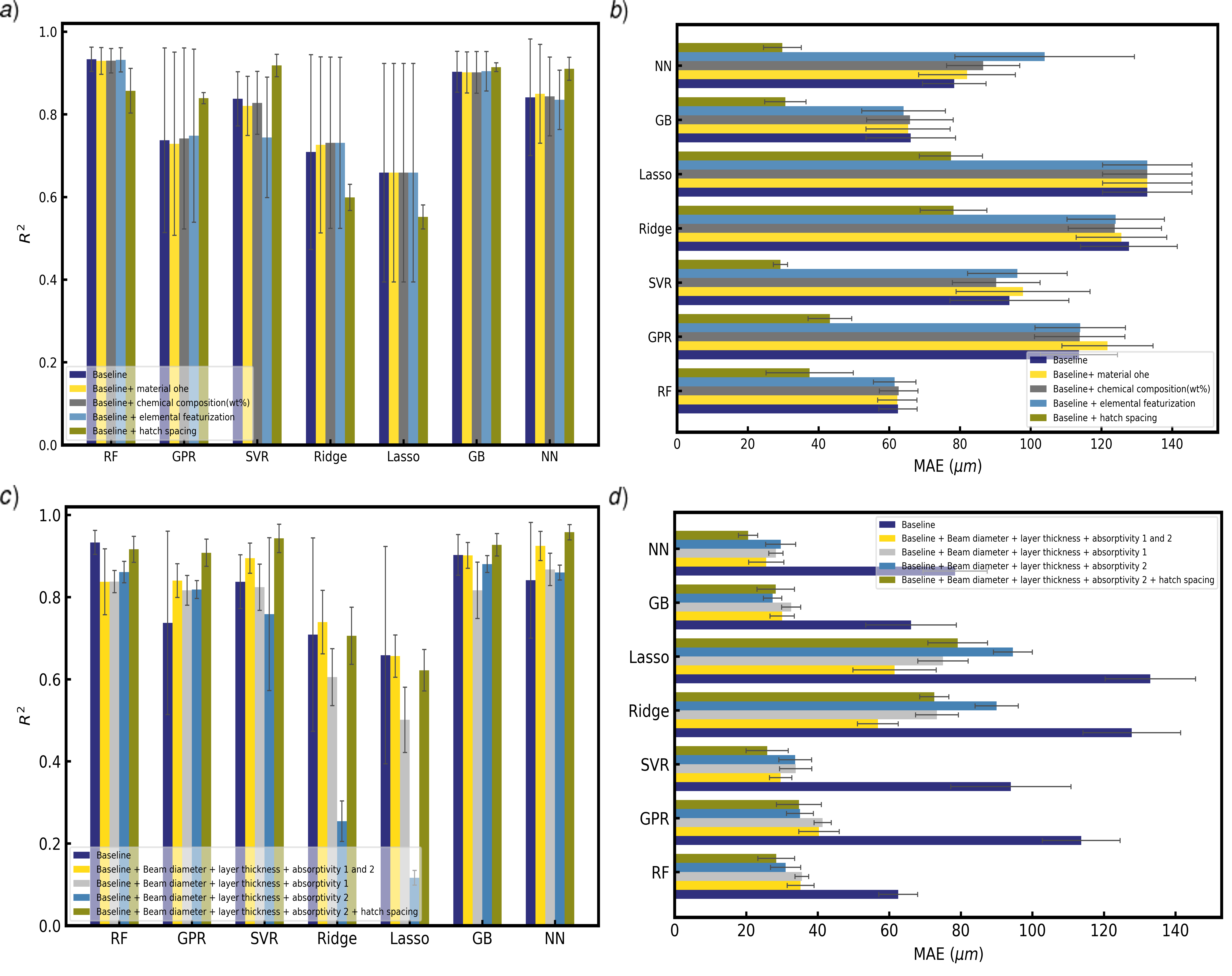}
\caption{Benchmark performances for depth of the melt pool, 'Random Forest', 'Gaussian Process Regressor','Support Vector Regressor', 'Ridge Linear Regressor', 'Lasso Linear Regressor', Gradient Boosting', 'Neural Network' models are evaluated by $R^2$ accuracy and MAE: a,c) A comparison between accuracy and MAE results of our models with different featurization. b,d) A comparison between accuracy and MAE results of our models on different combination of absorption coefficients as features. (Note: for $R^2$ and MAE, higher values and lower values indicate better performance, respectively.)}
\label{fig:fig4}
\end{figure}

\section{Results and Discussion}
In this section, the performance of the benchmarked models on datasets will be discussed. Note that all benchmark results presented in this paper are the average of five runs, with standard deviations listed as error bars. We also run a set of experiments focusing on how changing the size of training set affect model performance. Additionally, a Random Forest feature importance study was conducted in order to determine which features are most vital in each dataset and task. Details will be discussed in the following sections.
\vspace{3mm}

\subsection{\normalfont Task 1: Regression}

Accuracy ($R^2$) and MAE results of seven different machine learning models, i.e., 'Random Forest Regressor’, ’Gaussian Process Regressor’,’Support Vector Regressor’, ’Ridge Linear Regressor’, ’Lasso Linear Regressor’, ’Gradient Boosted Random Forest’, and ’Neural Network’ on the melt pool depth prediction have been reported (Fig. \ref{fig:fig4}).  We first run the ML models with the baseline features listed in Table \ref{tab:table2} (Fig.\ref{fig:fig4} baseline mode).  Although our data in hatch spacing was limited, adding hatch spacing as a new feature to the baseline featurization decreased the MAE noticeably (Fig.\ref{fig:fig4} baseline + hatch spacing model). 
\vspace{3mm}

To featurize the different alloys in the dataset, various material featurization methods were employed. On the first attempt, alloy density, thermal conductivity, specific heat, and melting temperature were added as features in the baseline featurization.(Fig.\ref{fig:fig4} baseline model) Afterward, material one hot encoding featurization (Fig.\ref{fig:fig4} baseline + material ohe model), and adding material chemical composition (wt\%) (Fig.\ref{fig:fig4} baseline + chemical composition mass percentage model) were tested. Finally, we did elemental featurization on each alloy in the datasets. (Fig.\ref{fig:fig4} baseline + elemental featurization model) 
\vspace{3mm}

The mean absolute error  for each featurization mentioned above is depicted in Fig. \ref{fig:fig4}c. As compared to the baseline model, adding hatch spacing increased the accuracy and decreased MAE. The best accuracy belongs to baseline featurization for Random forest algorithm with 93.32\% accuracy, and the minimum MAE belongs to baseline plus hatch spacing as features for Neural Network algorithm with an MAE of 29.80 $\mu m$. Other featurizations did not change the results noticeably. (Fig. \ref{fig:fig4}a and Fig. \ref{fig:fig4}c)

\begin{figure}\centering
\includegraphics[scale=0.093]{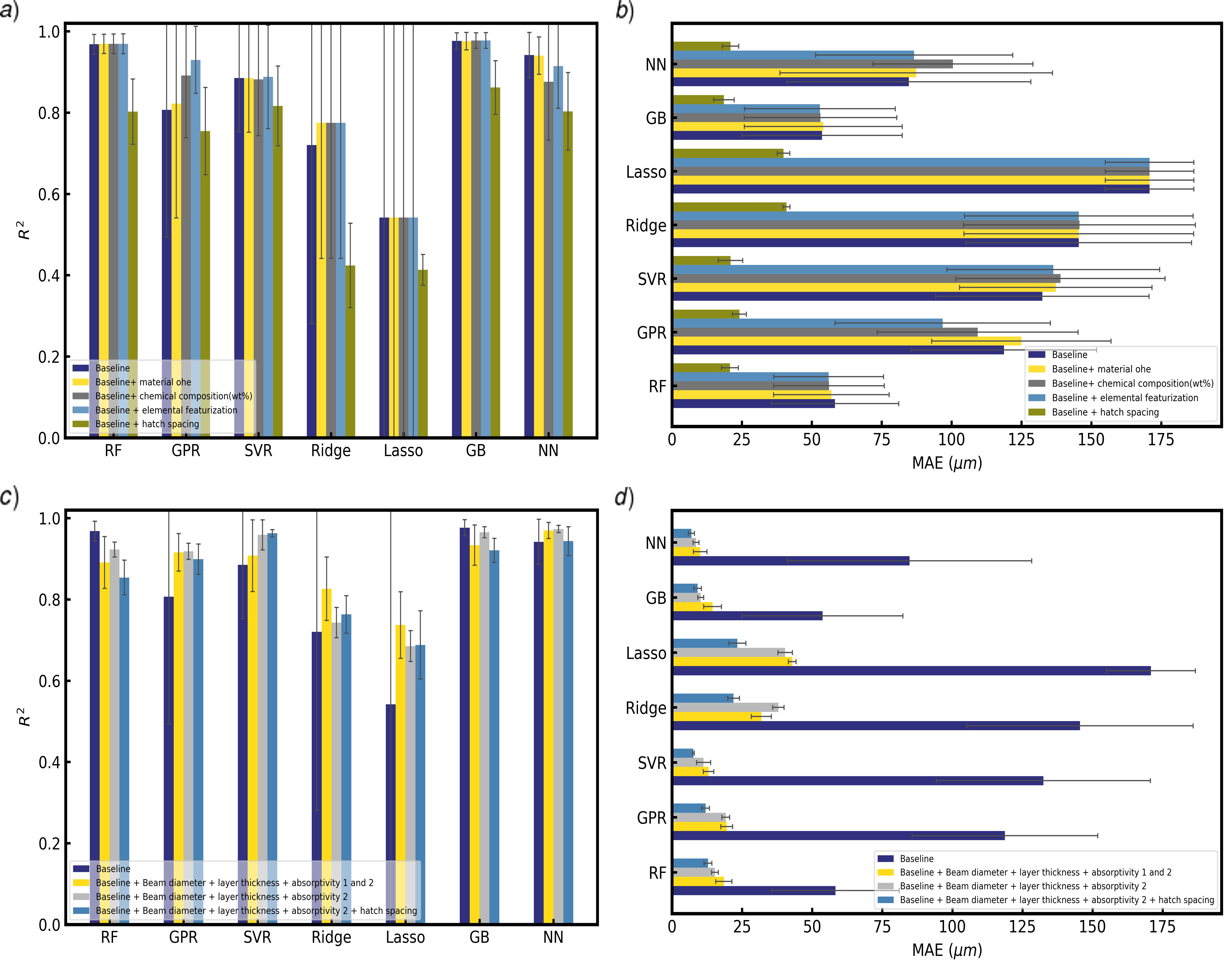}
\caption{Benchmark performances for width of the melt pool, 'Random Forest', 'Gaussian Process Regressor','Support Vector Regressor', 'Ridge Linear Regressor', 'Lasso Linear Regressor', Gradient Boosting', 'Neural Network' models are evaluated by $R^2$ accuracy and MAE: a,c) A comparison between accuracy and MAE results of our models with different featurization. b,d) A comparison between accuracy and MAE results of our models on different combination of absorption coefficients as features. (Note: for $R^2$ and MAE, higher values and lower values indicate better performance, respectively.)}
\label{fig:fig5}
\end{figure}

\vspace{3mm}

In addition, the absorptivity coefficients, explained in previous section, plus beam diameter and powder layer thickness were included as new features on the L-PBF dataset. As shown in Fig. \ref{fig:fig4}b, in comparison with baseline model, adding one or both of the empirical absorptivity coefficients boosted the results significantly. As the best result observed thus far was for a set of features that included hatch spacing, it was appended to the absorption coefficients which improved performance. Fig. \ref{fig:fig4}d represents the mean absolute error for the featurization discussed above. The best result for $R^2$ accuracy belongs to the baseline feature set with the second absorptivity coefficient and hatch spacing as our features, using the Neural Network model. This yields an accuracy of  95.80\% accuracy and an MAE of 20.54 $\mu m$. (Fig. \ref{fig:fig4}b and Fig. \ref{fig:fig4}d)

\vspace{3mm}


The accuracy ($R^2$) and MAE results of seven different ML models for the melt pool width prediction task has been illustrated (Fig. \ref{fig:fig5}). We first run ML models with baseline features. Similar to the regression model acting on the melt pool depth, adding hatch spacing as a new feature to the baseline featurization reduced the accuracy and MAE. The material one hot encoding, chemical composition (wt\%), and elemental featurization were also applied in order to observe their effects on the prediction accuracy (Fig. \ref{fig:fig4}a). The MAE for each featurization mentioned above is presented in Fig. \ref{fig:fig4}c. The best result is achieved with the Gradient boosting model, which yields 97.73\% accuracy for baseline plus elemental featurization and an MAE of 18.56 $\mu m$ for baseline plus hatch spacing (Fig. \ref{fig:fig5}a and Fig. \ref{fig:fig5}c). 

\vspace{3mm}

In addition, the absorptivity coefficients plus beam diameter and powder layer thickness were included as new features for the melt pool width prediction task on the LPBF dataset. As demonstrated in Fig. \ref{fig:fig5}b, adding the second absorptivity coefficient improved the results significantly (Fig. \ref{fig:fig5}d).  The best result for $R^2$ accuracy is achieved with the baseline featurization with second absorptivity coefficients added as our features for the Neural Network algorithm. With this configuration, the model achieves 97.32\% accuracy, and the minimum MAE of 6.92 $\mu m$ is observed with the baseline featurization and second absorptivity coefficient and hatch spacing (Fig. \ref{fig:fig5}b and Fig. \ref{fig:fig5}d). 

\vspace{3mm}

\begin{figure}\centering
\includegraphics[scale=0.093]{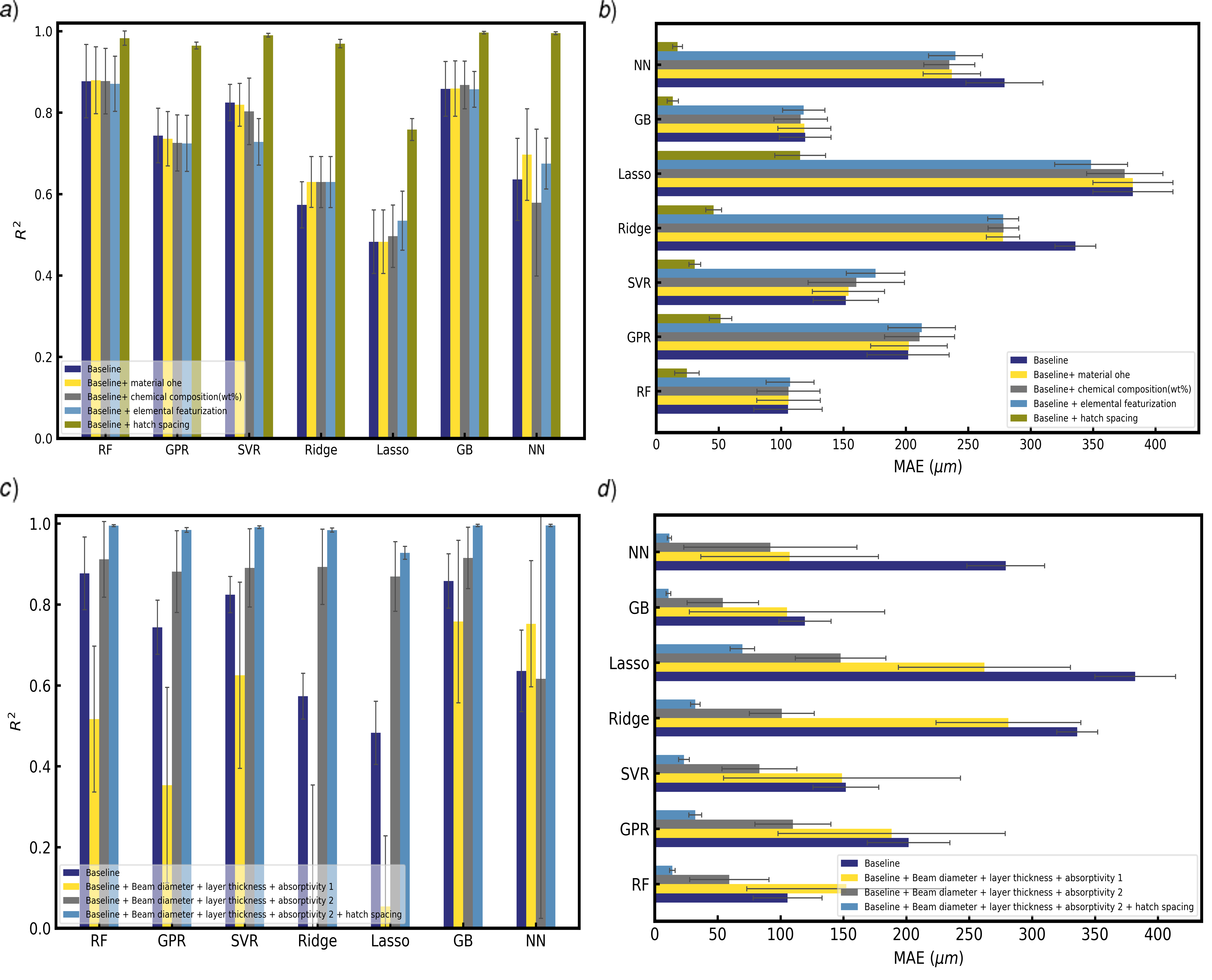}
\caption{Benchmark performances for length of the melt pool, 'Random Forest', 'Gaussian Process Regressor','Support Vector Regressor', 'Ridge Linear Regressor', 'Lasso Linear Regressor', Gradient Boosting', 'Neural Network' models are evaluated by $R^2$ accuracy and MAE: a,c) A comparison between accuracy and MAE results of our models with different featurization. b,d) A comparison between accuracy and MAE results of our models on different combination of absorption coefficients as features. (Note: for $R^2$ and MAE, higher values and lower values indicate better performance, respectively.)}
\label{fig:fig6}
\end{figure}

We have also assessed the accuracy ($R^2$) and MAE results of the ML models on the melt pool length prediction task (Fig. \ref{fig:fig6}).  Unlike the depth and width regression results, adding hatch spacing as a new feature to the baseline featurization increased the accuracy noticeably. Material one hot encoding, chemical composition (wt\%), and elemental featurization were also included (Fig. \ref{fig:fig6}a). These featurization models have been evaluated by mean absolute error as well (Fig. \ref{fig:fig6}c).  When compared with baseline model, adding hatch spacing raised the accuracy and decreased the MAE of the melt pool length prediction. The highest accuracy is produced by a model that incorporates the baseline features and hatch spacing, with 99.62\% accuracy and an MAE of 13.25 $\mu m$ for the Gradient boosting model. Other featurization results are presented in Fig. \ref{fig:fig6}a and Fig. \ref{fig:fig6}c. 

\vspace{3mm}

In addition, the absorptivity coefficients, along with the beam diameter and layer thickness were included as new features on the LPBF dataset. As presented in Fig. \ref{fig:fig6}b, in comparison with baseline model, even though adding the first absorptivity coefficient reduced the accuracies, the second absorptivity coefficient  boosted the results significantly. Since the best result so far was with hatch spacing, it was appended to absorptivity 2, and boosted the performance of the model (Fig. \ref{fig:fig6}c), and the mean absolute error for featurization discussed previously is reported in Fig. \ref{fig:fig6}d. The best result for $R^2$ accuracy is achieved with a model configuration of the baseline features, the second absorptivity coefficient and the hatch spacing as features for the Gradient boosting algorithms. Using the Gradient Boosting model, an accuracy of  99.55\% and an MAE of 10.92 $\mu m$ is achieved (Fig. \ref{fig:fig6}b and Fig. \ref{fig:fig6}d).

\vspace{3mm}

\begin{figure}\centering
\includegraphics[scale=0.087]{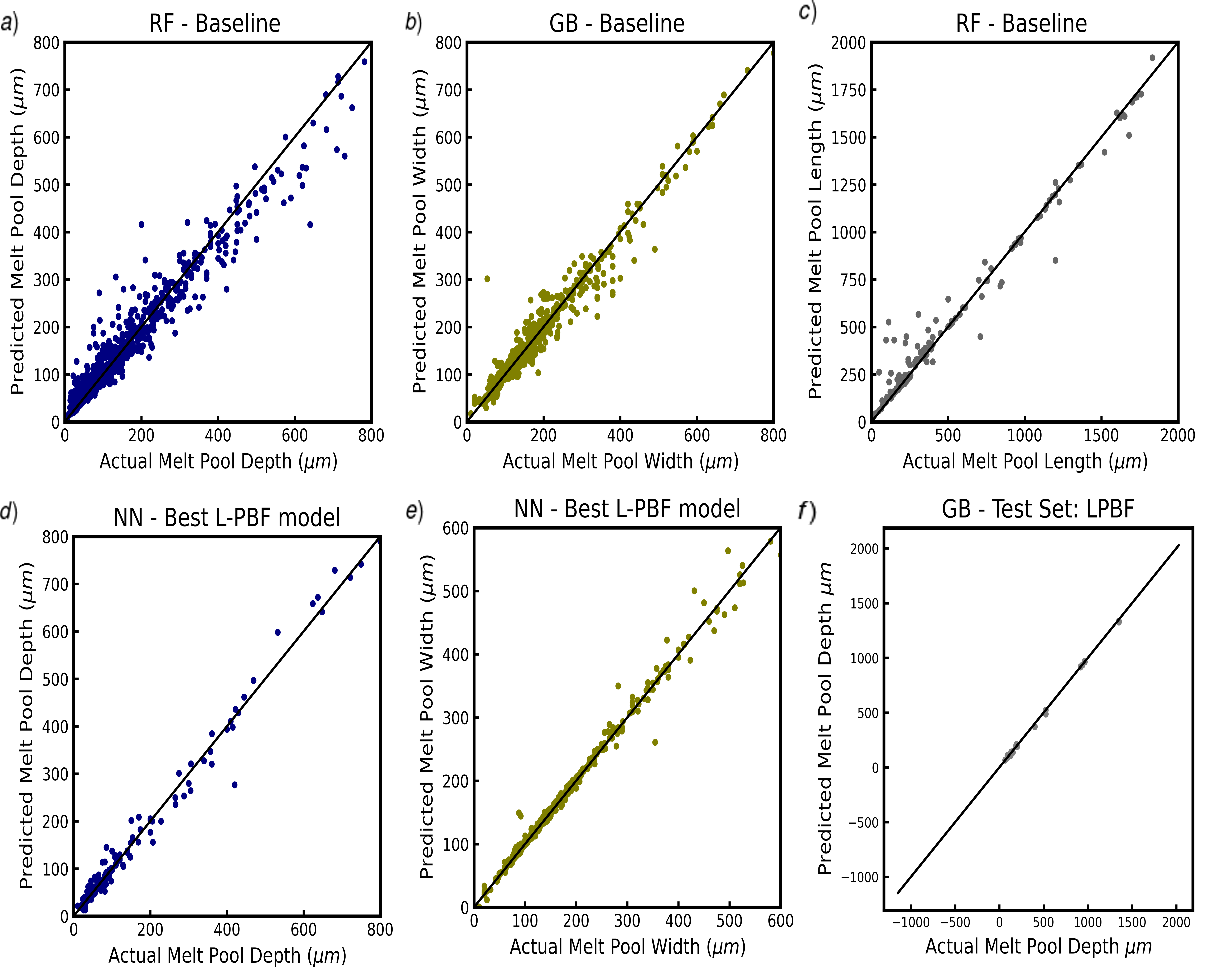}
\caption{Predicted vs. actual plots for melt pool depth, width, and length. a,b,c) For Random forest algorithm with baseline featurization. d,e,f) For Neural network algorithm with baseline plus absorptivities.}
\label{fig:fig7}
\end{figure}

A comparison between the predicted and actual values of the melt pool geometry in the test set partition is shown in  Fig. \ref{fig:fig7} for both the baseline featurization, and the best LPBF model. For a perfect fit, when a model has an $R^2$ of 1, all the points would be on the diagonal line, $y=x$. Thus, the results for the neural network model with baseline features plus absorptivities as features are thoroughly reliable, and predict accurate values to the actual geometries.

\vspace{3mm}

\begin{figure}\centering
\includegraphics[scale=0.078]{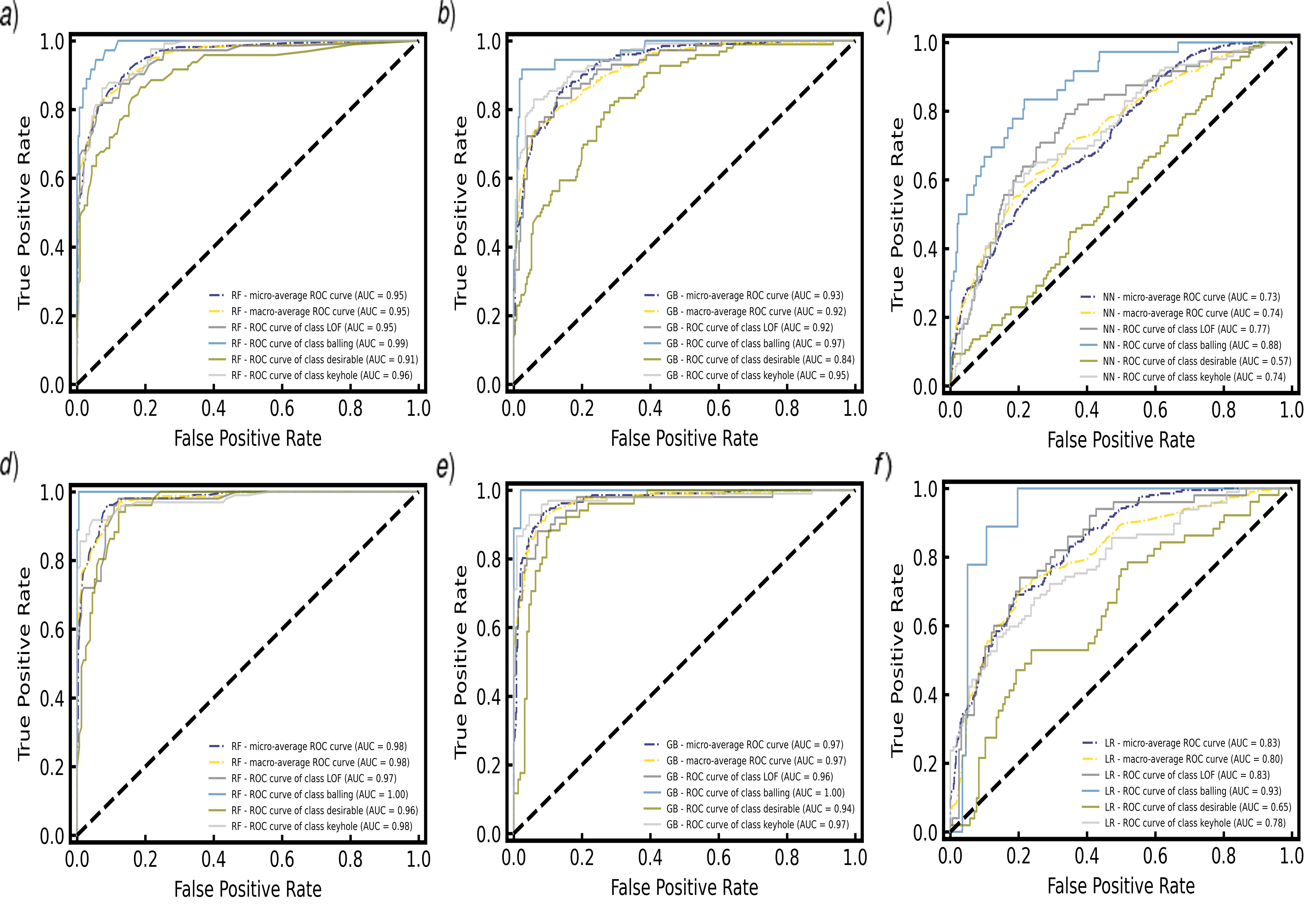}
\caption{Receiver operating characteristic (ROC) curves for prediction of melt pool classification of 'Random forest', Gradient Boosting' and 'Neural Network' models for 2 different featurization: a,b,c) baseline featurization. d,e,f) baseline plus absorption coefficient 1 as features. }
\label{fig:fig8}
\end{figure}

\begin{figure}\centering
\includegraphics[scale=0.09]{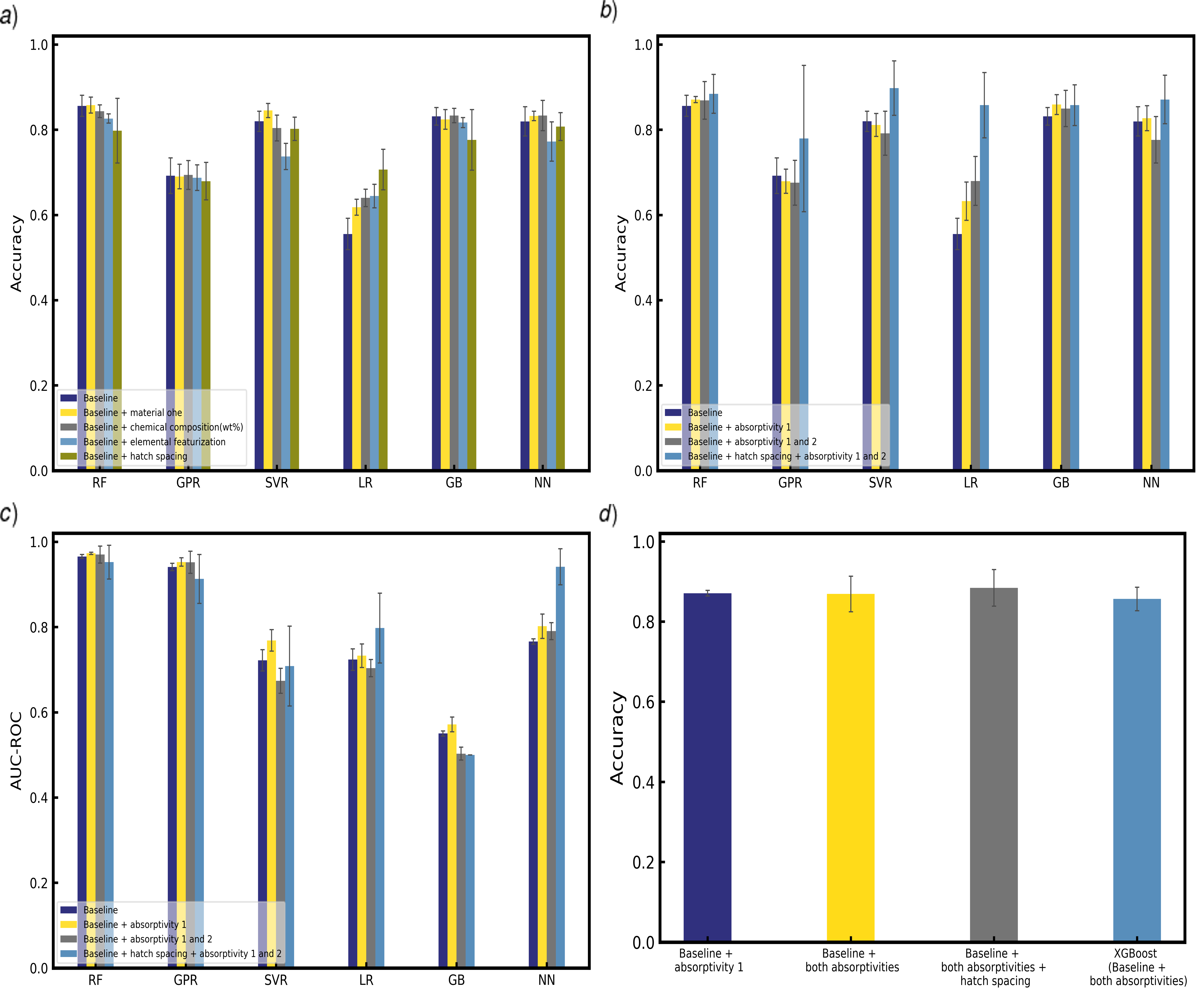}
\caption{Benchmark performances for melt pool classification, 'Random Forest', 'Gaussian Process Classifier','Support Vector Classifier', 'Logistic Regression', Gradient Boosting', 'Neural Network' models are evaluated by accuracy and AUC-ROC: a) Accuracy results of our models with different featurization. b) Accuracy results of our models on different combination of absorption coefficients as features. c) AUC-ROC results of our models with four different featurizations. (Note: for accuracy and AUC-ROC, higher values indicate better performances.) d) Comparing the accuracy of the other models with different input parameters compared to the model created by XGBoost.}
\label{fig:fig9}
\end{figure}

\subsection{\normalfont Task 2: Classification}

For the classification task, we used and extended the Receiver Operating Characteristic (ROC) metric to evaluate the prediction accuracy of the classifier output. In ROC curves, the X axis represents the false positive rate, and the Y-axis represents the true positive rate. This means that the larger the area under the curve (AUC), the better the model performs. Because ROC curves are typically utilized for binary classification, it is necessary to binarize the output in order to extend the ROC curve and ROC area to multi-label classification. For multi-class classification, we introduced two methods to evaluate every class at the same time: macro-average and micro-average. A macro-average will compute the metric independently for each class and then take the average, whereas a micro-average will aggregate the contributions of all classes to compute the average metric. We also included the ROC curve for each class in the result. The ROC curves of the best classifiers -- Random Forest, Gradient Boosting, Neural Network, and Logistic regressor -- have been shown for both the baseline featurization and the baseline with the first absorptivity coefficient included in Fig. \ref{fig:fig8}.

\vspace{3mm}

The accuracy and AUC-ROC results of six different ML algorithms, i.e., 'Random Forest’, ’Gaussian Process Classifier’,’Support Vector Classifier’, ’Logistic Regression’, ’Gradient Boosting’, ’Neural Network’ on melt pool LPBF classification prediction have been investigated. (Fig. \ref{fig:fig9}) Initially, we run the ML models with baseline features. Adding hatch spacing as a new feature to the baseline featurization boosted the accuracy to some extent. Material one hot encoding, chemical composition (wt\%), and elemental featurization were also applied (Fig. \ref{fig:fig9} a). The best result is achieved with a model configuration of the baseline features along with material one hot encoding as the input to the Random Forest algorithm with 85.78\% accuracy (Fig. \ref{fig:fig9} a).

\vspace{3mm}

Next, the absorptivity coefficients explained in previous section were included as new features. As shown in Fig. \ref{fig:fig9}b, as compared with baseline model, adding absorptivity coefficients improved the results significantly. The hatch-spacing feature was also appended to absorption coefficients, resulting in a better result for some ML models. The  AUC-ROC results for classification task are analyzed and shown in Fig. \ref{fig:fig9} c. An accurate result is achieved for this task using the baseline features along with both absorptivity coefficients and hatch spacing as input to the Random forest model. This results in an  accuracy of 88.42\% and an AUC-ROC metric of 0.98 (Fig. \ref{fig:fig9} b and Fig. \ref{fig:fig9} c). In addition, utilizing the XGBoost model in order to predict the melt pool classification including the parameters of both the absorption coefficients, baseline and elemental featurization, results in the accuracy of the classification task to 85.6\% as seen in Fig. \ref{fig:fig9} d.

\begin{figure}\centering
\includegraphics[scale=0.077]{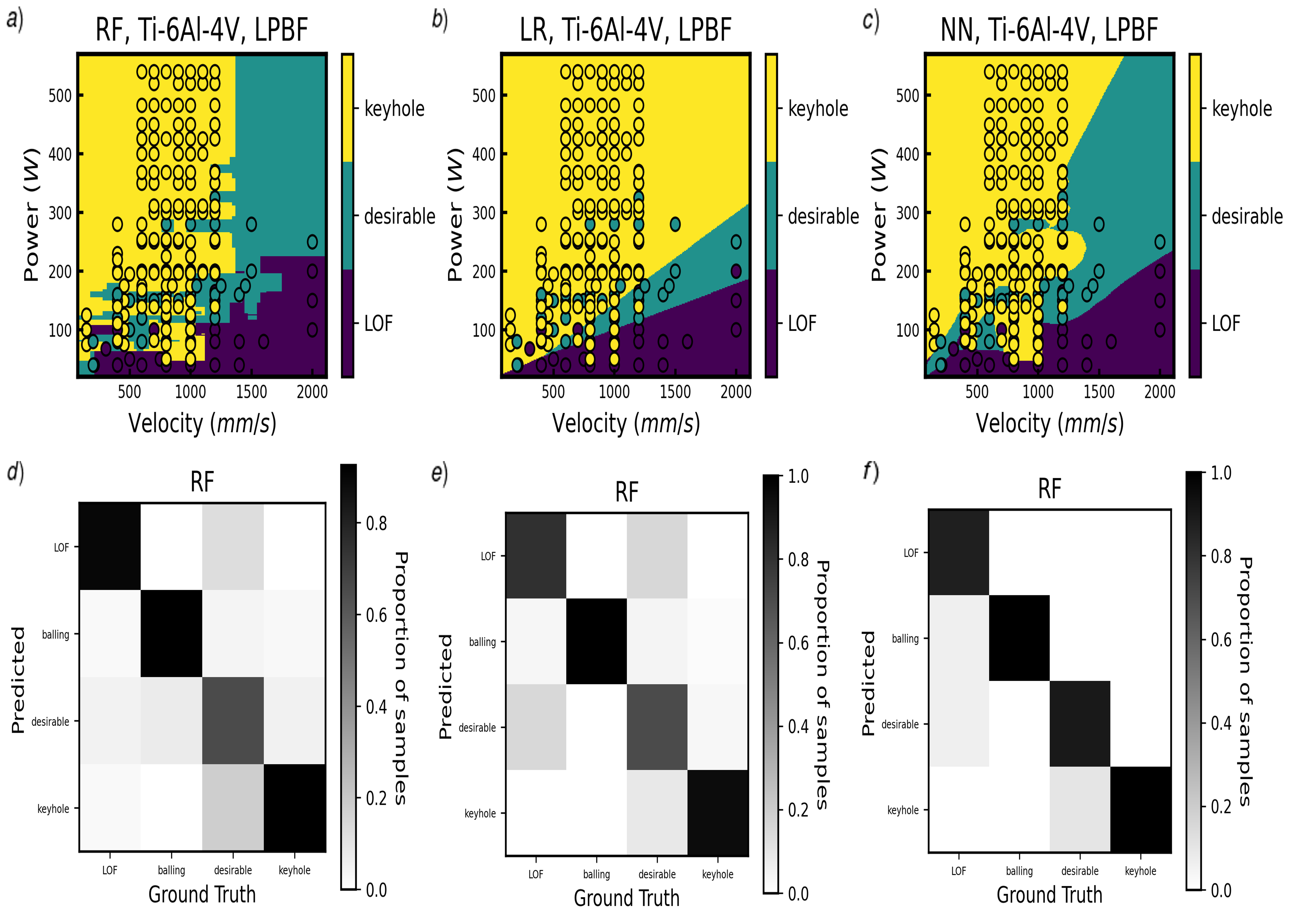}
\caption{Classification decision boundaries of our dataset based on power and velocity for Ti6-Al4-V alloy in PBF process a) for Random forest model, b) for Neural network, c) for Gradient boosting. Confusion matrix for classification task with d) with baseline plus absorptivity coefficient 1 as features e) with baseline plus absorptivity coefficient 2 f) with baseline plus both absorptivity coefficients and hatch spacing. }
\label{fig:fig10}
\end{figure}

\vspace{3mm}

Visualizing the decision boundaries of the classifiers allows us to examine their partition behavior. Therefore, the classification decision boundaries of the dataset based on power and velocity for Ti-6Al-4V in the LPBF process for Random Forest, Neural Network, and Logistic Regression models have been explored. (Fig. \ref{fig:fig10} a,b, ad c)

\vspace{3mm}

In addition, we also plot confusion matrices to visualize the performance of the implemented models. Each row of the matrix represents the occurrence of a class based on the model predictions, while each column represents the occurrence found in the ground truth. The confusion matrix for the best classifier, Random Forest, for baseline features plus absorption coefficient 1 is depicted in Fig. \ref{fig:fig10} d. Additionally, the confusion matrix for the best classifier, Random Forest, for baseline features plus both absorption coefficients, and for the combination of baseline features, absorption coefficients, and hatch spacing as features is shown in  Fig. \ref{fig:fig10} e and f. 
 
\subsection{\normalfont Size of dataset vs. accuracy}

\begin{figure}\centering
\includegraphics[scale=0.2]{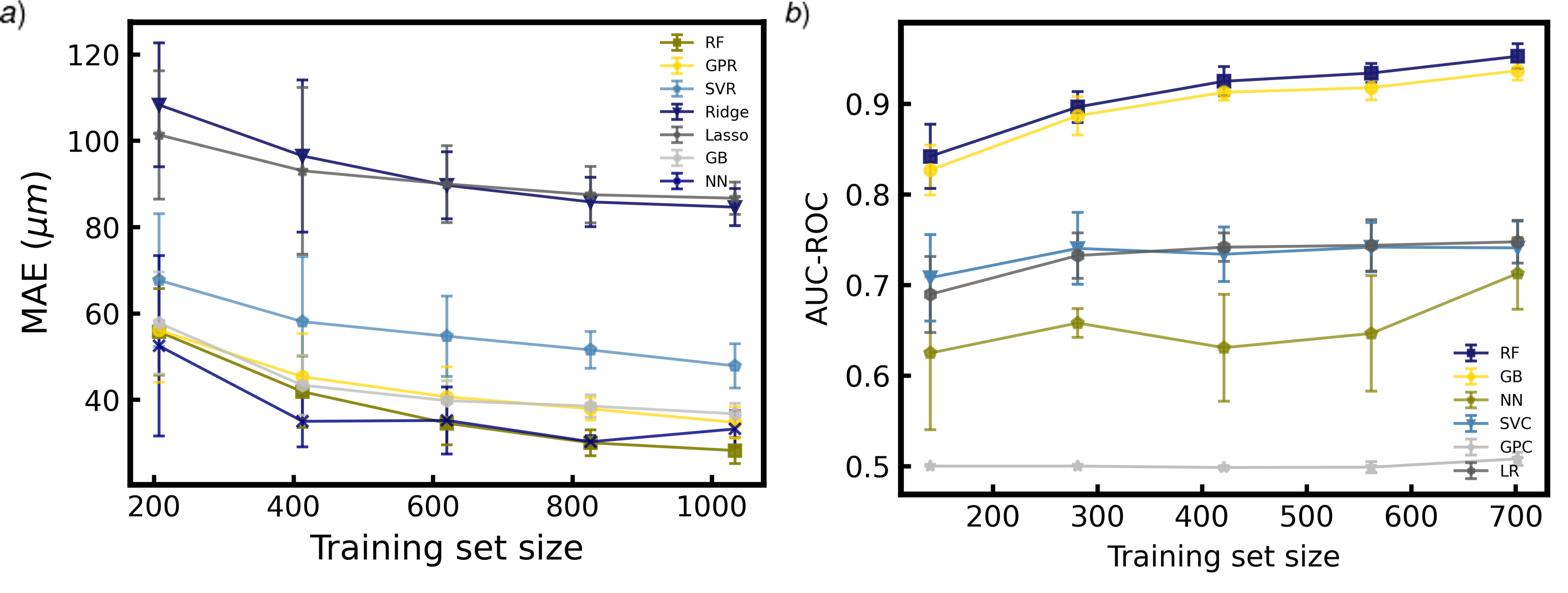}
\caption{Out-of-sample performances with different training set sizes with baseline featurization on a) melt pool depth b) melt pool classification. (Each data point is the average of 5 independent runs, with standard deviations shown as error bars.) }
\label{fig:fig11}
\end{figure}

Here, we present an investigation on how the performance of regression tasks changes with the increasing volume of data being sampled. We randomly selected 20\%, 40\%, 60\%, 80\%, 100\% of samples from the training section of the regression dataset, and then performed cross-validation for the different regressor models used. The mean and standard deviations of the test MAE from five independent runs following cross-validation for each regressor as a function of sample size are displayed in Fig. \ref{fig:fig11} a. We perform the baseline featurization model predicting melt pool depth. We can see the clear decrease of the test MAE as the sample size increases. In addition, we can conclude that the Random Forest and Neural Network performs better compared to other models.

\vspace{3mm}

Additionally, to illustrate how classification task performance changes with increasing sample size, we conduct trials with varying dataset sizes. We randomly selected 20\%, 40\%, 60\%, 80\%, 100\% of samples from the training dataset, and performed cross-validation for different classifiers. For baseline featurization model, the mean and standard deviations of AUC-ROC from five independent runs in cross-validation for each classifier as a function of training sample size is defined. (Fig. \ref{fig:fig11} b) A trend of increasing performance is observed for each classifier as we increase the sample size. For Random Forest and Gradient Boosting, we observe a clearer increase compare to other classifiers.

\vspace{3mm}

\begin{figure}\centering
\includegraphics[scale=0.115]{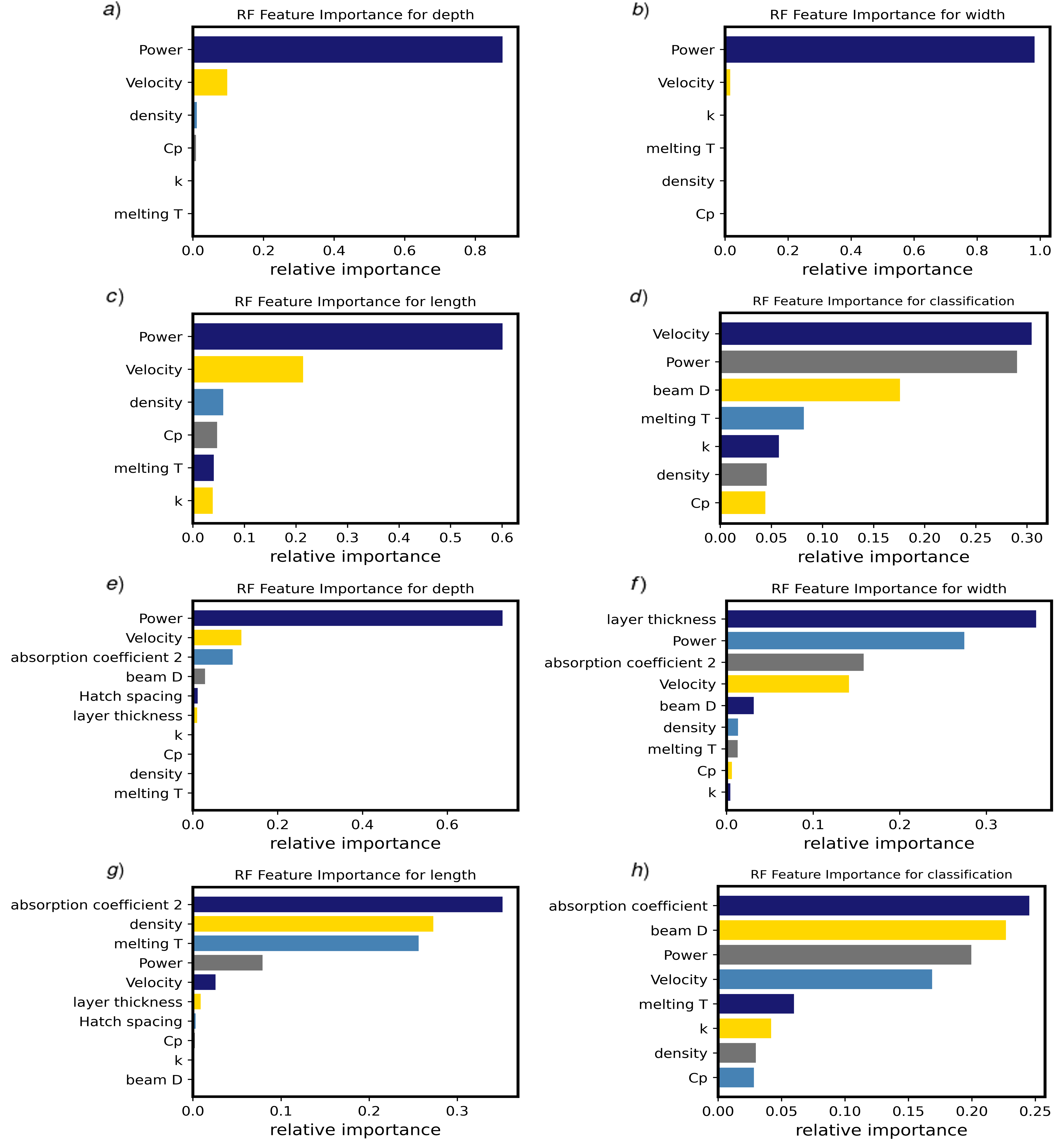}
\caption{Random forest feature importance with different featurization for melt pool geometry and classification: (a,b,c,d) With baseline featurization:( e, f, g)  With baseline plus absorptivities (h)}
\label{fig:fig12}
\end{figure}

\vspace{3mm}
 
\subsection{\normalfont Random forest feature importance}

In order to determine which features contribute the most towards the prediction performance of our model, we perform  feature importance study. The results of a random forest feature importance study, with baseline features with and without absorptivity features is presented in Fig. \ref{fig:fig12}. For the baseline featurization, beam power and velocity are shown to have a high importance followed by material thermal properties. (Fig. \ref{fig:fig12} a,b,c, and d) For the LPBF dataset with baseline featurization, absorptivities, beam diameter, and layer thickness, absorption coefficients have also ranked high, and are among the most important and determinant features. However, power, velocity, and beam diameter are related to how much energy is emitted by the heat source, and absorption coefficients and layer thickness are associated with the energy absorbed. Thus, power, velocity, layer thickness, beam diameter, and absorption coefficients all play a pivotal role in determining the input energy density and the energy absorbed.
\begin{table}[ht]
\begin{center}
\caption{Summary of machine learning models performances on regression task (depth, width, and length of the melt pool) and classification task (test subset). (Best performance value is reported in two different metrics for each category.)}\vspace{1mm}
\label{tab:table5}
\scalebox{0.7}{
\begin{tabular}{c| c c  c c c}
\toprule [1pt]
Category & Metric            & Best performance  & Value  \\
\midrule [2pt]
Depth of melt pool  & $R^2$ &  Neural Network  & 0.9580 \\ [1ex]
  & MAE &  Neural Network & 20.54 \\ [1ex]

Width of melt pool  & $R^2$ & Gradient boosting & 0.9773\\ [1ex]
  & MAE  & Neural Network & 6.92 \\ [1ex]

Length of melt pool  & $R^2$ & Gradient Boosting  & 0.9962 \\ [1ex]
  & MAE  & Gradient Boosting  & 10.92 \\ [1ex]

Melt pool classification & Classification accuracy  & Random Forest & 0.8841 \\ [1ex]
 & AUC-ROC & Random Forest & 0.98\\ [1ex]

\bottomrule
\end{tabular} 
}
\end{center}
\end{table} 

\subsection{\normalfont Model Suitability}

 The models that are best suited to datasets with complex decision boundaries that are informed by a large number of features are Random Forest ensemble methods (RF, GB and XGBoost) and deep models such as Neural Networks (Table \ref{tab:table5}). This is demonstrated by our results, as these two classes of algorithms consistently outperform the other algorithms explored. While kernel-based models, such as SVMs, show decent performance on the dataset, they are better suited for binary classification problems, as opposed to the multiclass classification performed here \cite{crammer2001algorithmic}. Gaussian Process methods are suited to non-linear decision boundaries, but are less feasible in high-dimensional settings due to their computational expense \cite{liu2020gaussian}. Finally, the linear methods explored in this work - Lasso, Ridge, and Logistic Regression  - are unable to effectively characterize the complex decision boundaries required, due to limitations in their representation capabilities.
 
 \vspace{3mm} 
In addition, Random Forest models are suited to data with heterogeneous scales, and can readily combine categorical and numerical features. Therefore, Random Forest-based models and Neural Network models are more robust and scalable for a dataset that may be expanding in an ongoing manner. 


\subsection{Model Identification}

Machine learning algorithms provide effective predictive models for different tasks; however, many of these methods cannot be easily interpreted. Interpretability is of particular interest in scientific applications where the systems are usually governed by some underlying equations. To this end, we try to extract explicit relations between the process and material properties from dataset. 
\vspace{3mm} 
The literature of equation discovery methods consists of symbolic regression techniques \cite{ai-feynman} and parsimonious models with sparse regression \cite{Sindy}. However, the large number of parameters in the AM process and the noisy data renders it unfeasible for the current symbolic regression methods that are mainly studied for cases with single covariate. Also, the complexity of the process and the possible rational forms of equations makes this identification a difficult task for sparse regression frameworks. Therefore, we construct a tailored identification framework which can handle several parameters under some prior assumptions about the model. 
\vspace{3mm} 
One of the main conditions in physical equations is dimension analysis. Based on dimension analysis, given a complete set of parameters with their dimension, the dimension of units should be consistent on both sides of an equation. We can easily conclude that a simple linear regression cannot cover the relations between different parameters that have various dimensions. Therefore, we consider a nonlinear regression model as a constrained optimization problem where the constraints are the dimension consistency. 
\vspace{3mm} 
Similar to the regression tasks, we aim to find a predictive model for the melt pool geometry including its depth, width, and length. While many parameters can affect these properties, we limit our search to a specific set of material properties and process conditions. For each task, a subset of the dataset without missing values corresponding to the covariates and the labels are selected. The parameters and their dimensions are tabulated in table \ref{table:dim}.

\begin{table}[ht]
\centering
\caption{List of parameters and their dimensions}
\begin{tabular}{c|c|c c c c}
\toprule [1pt]
Parameters & Unit (SI) &$M$ & $L$ & $T$ & $K$ \\
\midrule [2pt]
D, W, L & m &0 & 1 & 0 & 0 \\
\midrule [1pt]
P & W & 1 & 2 & -3 & 0 \\[3pt]
V & $\frac{m}{s}$ & 0 & 1 & -1 & 0  \\[3pt]
$\rho$ & $\frac{kg}{m^3}$ & 1 & -3 & 0 & 0 \\[3pt]
$C_p$ & $\frac{m^2}{s^2 K}$ & 0 & 2 & -2 & -1\\[3pt]
$k$ & $\frac{kg m}{s^3 K}$ & 1 & 1 & -3 & -1\\[3pt]
$T_m$ & K& 0 & 0 & 0 & 1\\
\bottomrule
\end{tabular}
\label{table:dim}
\end{table}

The nonlinear form we consider is a multiplication of parameters with a constant variable as a multiplier ($w_0$), and one exponent variable for each parameter (Eq. \ref{eq:nonlinear-form}). Four constraints correspond to the satisfaction of dimension analysis for each basic units of mass, length, time, and temperature based on table \ref{table:dim}. The optimization problem is derived by considering the $L_2$ norm of the equation as an objective function to minimize, and compacting the constraints into a single equality constraint as in (Eq. \ref{eq:optimization}). 

\begin{equation}
\begin{aligned}
 y = & w_0 P^{w_1} V^{w_2} \rho^{w_3} C_p^{w_4} k^{w_5} (T_m - T_0)^{w_6} \\
\textrm{where}: \quad & 1) w_1 +w_3 +w_5 = 0 \\
& 2) 2w_1 + w_2 - 3w_3 + 2w_4 + w_5 = 1 \\
& 3) -3w_1 - w_2 - 2w_4 - 3w_5 = 0 \\
& 4) -w_4 - w_5 + w_6 = 0
\end{aligned}
\label{eq:nonlinear-form}
\end{equation}

\begin{equation}
\begin{aligned}
\min_{W} \quad & \| y - w_0 P^{w_1} V^{w_2} \rho^{w_3} C_p^{w-4} k^{w_5} (T_m - T_0)^{w_6} \|_2^2 \\
\textrm{s.t.}: \quad & (w_1 +w_3 +w_5)^2 + (2w_1 + w_2 - 3w_3 + 2w_4 + w_5 - 1)^2 +\\
& (-3w_1 - w_2 - 2w_4 - 3w_5)^2 + (-w_4 - w_5 + w_6)^2 = 0 \\
\end{aligned}
\label{eq:optimization}
\end{equation}

We use the SciPy \cite{2020SciPy-NMeth} constrained optimization package with trust region methods \cite{trust-region} to solve the optimization problem. The solutions are not guaranteed to reach global minimum and we report a sample solution for each of the tasks in table \ref{table:eq-results}

\renewcommand{\arraystretch}{1.5}
\begin{table}[ht]
\centering
\caption{Identified equations for modeling meltpool geometry and their $R^2$}
\scalebox{0.9}{
\begin{tabular}{c|c|c}
\toprule [1pt]
Task (label) & Equation &$R^2$ \\
\midrule [2pt]
Depth &  $D= 0.37 \times 10^6 \times P^{0.51} V^{-0.46} \rho^{-0.46} C_p^{-0.46} k^{-0.06} (T_m - T_0)^{-0.51}$ & 0.787 \\

Width & $W = 0.44 \times 10^6 \times P^{0.59} V^{-0.36} \rho^{-0.39} C_p^{-0.39} k^{-0.21} (T_m - T_0)^{-0.60}$ & 0.991   \\

Length & $L = 63.75 \times 10^6 \times P^{0.46} V^{-0.07} \rho^{-0.52} C_p^{-0.74} k^{0.06} (T_m - T_0)^{-0.68}$ & 0.473\\
\bottomrule
\end{tabular}
}
\label{table:eq-results}
\end{table}

One of the interesting observations arises when comparing the identified equations to the approximations of geometry derived from the theoretical Rosenthal equation \cite{10014566598}:

\begin{equation}
\begin{aligned}
(T - T_0) = \frac{Q}{2 \pi K R}\exp{\frac{\rho C_p V (Z-R)}{2 K}}
\end{aligned}
\label{eq:rosenthal}
\end{equation}

where $Q$ is the absorbed power, $V$ is the beam scanning speed, $K$ is the alloy thermal conductivity, $C_p$ is the alloy specific heat, $\rho$ is the alloy density, $T$ is temperature, $T_0$ the temperature far from the melt pool, $Z$ is horizontal distance from the moving point heat source (opposite to V direction), and $R$ is defined as $Z^2 + r^2$. Here, $r$ is the vertical distance from the point heat source line. Equations \ref{eq:rosenthal}a-c show the approximations of the melt pool depth, width, and length (by putting $Z=R$) derived from the Rosenthal equation, respectively \cite{TANG201739}.

\begin{subequations}
\begin{equation}
\begin{aligned}
& D_R = \sqrt{\frac{AQ}{e \pi \rho C_p (T_m - T_0) V}} 
\end{aligned}
\label{eq:rosenthal depth}
\end{equation}

\begin{equation}
\begin{aligned}
& W_R = 2 \times D_R 
\end{aligned}
\label{eq:rosenthal width}
\end{equation}

\begin{equation}
\begin{aligned}
& L_R = \frac{Q}{2 \pi K (T_m - T_0)}
\end{aligned}
\label{eq:rosenthal Length}
\end{equation}
\label{eq:rosenthal}
\end{subequations}

where $T_m$ is melting temperature, $e$ is the basis of natural logarithms, and $A=2.5$ for AlSi10Mg, and $A=2$ for other alloys. We can observe that the identified equation for the depth of meltpool in table \ref{table:eq-results} has quite a similar form to the Rosenthal estimated depth (equation \ref{eq:rosenthal depth}) which indicates the consistency of the equation found from the data and the underlying physics.

\begin{table}[ht]
\centering
\caption{$R^2$ accuracy comparison for Rosenthal, identified equations, and ML models for meltpool depth, width, and length}
\scalebox{0.9}{
\begin{tabular}{c|c|c|c}
\toprule [1pt]
Task (label) & Rosenthal $R^2$ & Identified equation $R^2$ & ML $R^2$\\
\midrule [2pt]
Depth &  0.654 & 0.787 & 0.958 \\

Width & 0.782 & 0.991 &  0.9773  \\

Length & -0.15 & 0.473 & 0.9962  \\
\bottomrule
\end{tabular}
}
\label{table:comparison}
\end{table}
\vspace{3mm} 
To compare the predictive performance of the different models, we report the $R^2$ of the Rosenthal estimations (equations \ref{eq:rosenthal}), constitutive equations from data (table \ref{table:eq-results}), and the best ML model together in table \ref{table:comparison}. Although ML models are not interpretable, they usually achieve the best predictive performance. The proposed constitutive models, however, obtain higher $R^2$ values than Rosenthal equations while keeping the explicit form that leads to their interpretability. 


\section{Conclusion}
This work introduces a comprehensive machine learning benchmark for melt pool geometry and defect type prediction. We collected data from a wide range of AM experiments performed on melt pool characterization. Various methods of feature engineering for AM input data were introduced to enhance the accuracy of different ML models. Multiple ML models were benchmarked against each other in conjunction with different schemes of featurization. Evaluation metrics and the standard reporting practices were also discussed in our work. For regression and classification, neural networks, Gradient boosting, and random forest outperforms the other ML models. In addition, we demonstrated that feature engineering in AM is important to yield  a highly accurate and generalizable ML model. Additionally, a data-driven model identification method was developed to estimate the meltpool geometry based on the dataset processing parameters and material properties. The proposed explicit models not only were more interpretable as compared to the employed ML models, but also showed better prediction performance for meltpool geometry as opposed to the Rosenthal meltpool geometry estimation. 
By providing a uniform platform for comparison and evaluation, we hope our benchmark facilitates the optimization and control of additive manufacturing processes and MeltpoolNet becomes a comprehensive resource for the metal additive manufacturing machine learning community.
\vspace{3mm}

\section*{Acknowledgements}
Research was sponsored by the Army Research Laboratory and was accomplished under Cooperative Agreement Number W911NF-20-2-0175. The views and conclusions contained in this document are those of the authors and should not be interpreted as representing the official policies, either expressed or implied, of the Army Research Laboratory or the U.S. Government. The U.S. Government is authorized to reproduce and distribute reprints for Government purposes notwithstanding any copyright notation herein.


\appendix

\section{Supplementary data}
\label{sec:sample:appendix}
The code and data associated with this article will be available online upon publication. 

\begin{table}[hbt!]
\begin{center}
\caption{Chemical composition (wt\%) of alloys studied in our benchmark.}\vspace{1mm}
\label{sec:sample:appendix:tab:tableA1}
\scalebox{0.6}{
\begin{tabular}{c c c c c c c c c c c c c c c c c c c c}
\toprule [1pt]
{alloys} & Y & Zn & Mg	& Si &	Al &Sn &	Zr &	W & Ti &	V &	Co &	Cu &	Ta	& Nb &	Ni &	Cr &	Fe &	Mn & Mo  \\ [1ex]
\midrule [2pt]
SS316L & 0&	0&	0&	0&	0&	0&	0&	0&	0&	0&	0&	0&	0&	0&	11.4&	17.3&	65&	1.5&	2.5 \\
Ti-6Al-4V & 0&	0&	0&	0&	5.5&	0&	0& 0& 90&	4.2&	0&	0&	0&	0&	0&	0&	0&	0&	0\\
IN718 & 0&	0&	0&	0&	0&	0&	0&	0&	1&	0&	0&	0&	0&	5.5&	55&	21&	0&	0&	3.3 \\
SS17-4PH & 0&	0&	0&	0&	0&	0&	0&	0&	0&	0&	0&	3.65&	0&	0&	4&	16&	75&	0&	0\\
IN625 & 0&	0&	0&	0&	0&	0&	0&	0&	0&	0&	0&	0&	0&	3.35&	65&	21&	0&	0&	9\\
IN738LC & 0&	0&	0&	0&	3.5&	0&	0&	2.6&	3.5&	0&	8&	0&	1.75&	1&	61&	15&	0&	0&	0\\
Hastelloy X & 0&	0&	0&	0&	0&	0&	0&	0&	0&	0&	0&	0&	0&	0&	45.5&	21.75&	18.5&	0&	9\\
Cu10Sn & 0&0&	0&	0&	0&	10&	0&	0&	0&	0&	0&	90&	0&	0&	0&	0&	0&	0&	0\\
AlSi10Mg & 0&	0&	0.35&	9.5&	90&	0&	0&	0&	0&	0&	0&	0&	0&	0&	0&	0&	0&	0&	0\\
Al-2.5Fe & 0&	0&	0&	0&	97&	0&	0&	0&	0&	0&	0&	0&	0&	0&	0&	0&	2.5&	0&	0\\
Tungsten & 0&	0&	0&	0&	0& 0&	0&	100&	0&	0&	0&	0&	0&	0&	0&	0&	0&	0&	0\\
Al-C-Co-Fe-Mn-Ni & 0&	0&	0&	0&	3&	0&	0&	0&	0&	0&	25&	0&	0&	0&	25&	0&	23.6&	23.2&	0\\
Ti-49Al-2Cr-2Nb & 0	&0&	0&	0&	49&	0&	0&	0&	47&	0&	0&	0&	0&	2&	0&	2&	0&	0&	0\\
HCP Cu & 0&	0&	0&	0&	0&	0&	0&	0&	0&	0&	0&	100&	0& 0&	0&	0&	0&	0&	0\\
Invar36 & 0&	0&	0&	0&	0&	0&	0&	0&	0&	0&	0&	0&	0&	0&	36&	0&	64&	0&	0\\
SS304 & 0&	0&	0&	0&	0&	0&	0&	0&	0&	0&	0&	0&	0&	0&	8&	18&	70&	2&	0\\
WE43 & 4.3&	0&	93&	0&	0&	0&	0&	0&	0&	0&	0&	0&	0&	0&	0&	0&	0&	0&	0\\
MS1 & 0&	0&	0&	0&	0.15&	0&	0&	0&	0.8&	0&	9&	0&	0&	0&	18&	0&	65&	0&	5\\
CMSX-4 & 0&	0&	0&	0&	5.6&	0&	0&	6.4&	1&	0&	9.5&	0&	6.4&	0&	60&	6.4&	0&	0&	0\\
TiC/IN 718 & 0&	0&	0&	0&	0.3&	0&	0&	0&	0.9&	0&	0&	0&	0&	5.1&	52&	18.4&	17.7&	0&	4.2\\
SS304L & 0&	0&	0&	0&	0&	0&	0&	0&	0&	0&	0&	0&	0&	0&	8&	18&	70&	2&	0\\
Ti6242 & 0&	0&	0&	0&	6&	2&	4&	0&	82&	0&	0&	0&	0&	0&	0&	0&	0&	0&	2\\
Ti-45Al & 0&	0&	0&	0&	45&	0&	0&	0&	48&	0&	0&	0&	0&	5&	0&	2&	0&	0&	0\\
K403 superalloy & 0&	0&	0&	0&	0&	0&	0&	5.2&	2.1&	0&	5.4&	0&	0&	0&	70&	11.3&	0&	0&	4.2\\
4140 steel & 0&	0&	0&	0&	0&	0&	0&	0&	0&	0&	0&	0&	0&	0&	0&	1.1&	97&	0.9&	0.2\\
AA7075 & 0&	5.5&	2.5&	0&	90&	0&	0&	0&	0&	0&	0&	1.5&	0&	0&	0&	0.3&	0.1&	0.3&	0\\
Ni-5NB & 0& 0&	0&	0&	0&	0&	0&	0&	0&	0&	0&	0&	0&	5&	95&	0&	0&	0&	0\\
Co-Cr-Fe-Mn-Ni &0 & 0&	0&	0&	0&	0&	0&	0&	0&	0&	19.4&	0&	0&	0&	28&	14.2&	19&	19.4&	0\\
Zn-2Al & 0&	98&	0&	0&	2&	0&	0&	0&	0&	0&	0&	0&	0&	0&	0&	0&	0&	0&	0\\
\bottomrule 
\end{tabular} }
\end{center}
\end{table}

\begin{table}[hbt!]
\begin{center}
\caption{Thermal properties of alloys studied in our benchmark.}\vspace{1mm}
\label{sec:sample:appendix:tab:tableA2}
\scalebox{0.7}{
\begin{tabular}{c c c c c}
\toprule [1pt]
{Alloys} & Density ($\frac{kg}{m^3}$) & Specific heat ($\frac{J}{kg.K}$) & Thermal conductivity ($\frac{W}{m.K}$) & Melting temperature (K) \\ [1ex]
\midrule [2pt]
SS316L & 8000&	500&	16.3&	1688\\
Ti-6Al-4V & 4470.5&	561.5&	7.2&	1922\\
IN718 & 8190&	435&	11.2&	1673\\
SS17-4PH & 7750&	460&	17.9&	1693\\
IN625 & 8442&	429&	9.8&	1593\\
IN738LC & 8110&	558.3&	19.8&	1543\\
Hastelloy X & 8220&	486&	9.1&	1628\\
Cu10Sn & 8780&	377&	50&	1272\\
AlSi10Mg & 2590&	910&	110&	1142\\
Tungsten & 19250&	134&	164&	3695\\
Ti-49Al-2Cr-2Nb & 3900&	610&	10.5&	1795\\
HCP Cu & 8960&	384.6&	399&	1631\\
Invar36 & 8050&	515&	10.15&	2000\\
SS304 & 8030&	500&	16.2&	1693\\
WE43 & 1800&	966&	51.3&	863\\
MS1 & 8209&	450&	31.4&	2848\\
CMSX-4 & 8100&	386.5&	45&	1653\\
TiC/IN 718 & 7377.5&	407.25&	13.25&	2015\\
SS304L & 8030&	500&	16.2&	1693\\
Ti6242 & 4540&	460&	6.92&	1978\\
4140 steel & 7850&	473&	42.6&	1689\\
AA7075 & 2810&	960&	130&	829\\
Ni-5NB & 8900&	550&	85&	1703\\
Co-Cr-Fe-Mn-Ni &7700&	600&	20&	1644\\
\bottomrule 
\end{tabular} }
\end{center}
\end{table}

\begin{table}[hbt!]
\begin{center}
\caption{Hyperparameters and their range studied in our benchmark ML models for meltpool width and length prediction.}\vspace{1mm}
\label{sec:sample:appendix:tab:tableA3}
\scalebox{0.5}{
\begin{tabular}{c c c c c}
\toprule [1pt]
ML Task & {Models} & Hyperparameters & Value &  Range studied   \\ [1ex]
\midrule [2pt]
 Meltpool width Regression & RF & n\_estimators &	500 &1-500	\\ [2ex]
  & GPR & kernel & ConstantKernel(1.0, (1e-1, 1e3))*RBF(1.0, (1e-3, 1e3)	&		RBF, DotProduct, Matern, RationalQuadratic \\ [1ex]
 & SVR & C & 1000	&1-1000	\\ [0.8ex]
 &  & kernel & 'rbf'	&['linear', 'poly', 'rbf', 'sigmoid']	\\ [2ex]
 & GB &n\_estimators & 495 &1-500	\\ [2ex]
 & NN & number of neurons &(64,32,32)	&[32, 64, 128, 256, 512]\\ [0.8ex]
 &  & alpha &0.00361218	& 1e-7 -  1e-1\\ [2ex]
 Meltpool length Regression & RF & n\_estimators &	500 &1-500	\\ [2ex]
  & GPR & kernel &RationalQuadratic(alpha=1, length\_scale=1)	&	RBF, DotProduct, Matern, RationalQuadratic\\ [1ex]
 & SVR & C & 275	&1-1000	\\ [0.8ex]
 &  & kernel & 'rbf'	&['linear', 'poly', 'rbf', 'sigmoid']	\\ [2ex]
 & GB &n\_estimators & 353 &1-500	\\ [2ex]
 & NN & number of neurons &(512,512,512)	&[32, 64, 128, 256, 512]\\ [0.8ex]
 &  & alpha &0.02152036	& 1e-7 -  1e-1\\ [2ex]

\bottomrule 
\end{tabular}
}
\end{center}
\end{table}

\clearpage

\bibliographystyle{elsarticle-num} 
\bibliography{mybibliography} 

\end{document}